\documentclass[letterpaper, 10 pt, conference]{ieeeconf}  

\IEEEoverridecommandlockouts                              

\usepackage{graphics}
\usepackage{graphicx} 
\usepackage{times} 
\usepackage{amsmath} 
\usepackage{amssymb}  
\usepackage{subfigure}
\usepackage{color}
\usepackage{epsfig}
\usepackage{epstopdf}
\usepackage{etoolbox}
\usepackage{xcolor}
\usepackage{algorithm}
\usepackage{algpseudocode}
\usepackage{booktabs} 
\usepackage{cite}
\usepackage{subfigure}
\usepackage{balance}
\usepackage[font=footnotesize]{caption}

\DeclareMathOperator*{\argmax}{arg\,max}

\def\real{\mathbb{R}}

\title{\LARGE \bf
Cooperative Grasping for Collective Object Transport \\ in Constrained Environments
}

 \author{David Alvear, George Turkiyyah, and Shinkyu Park
 \thanks{The work was supported by funding from King Abdullah University
 of Science and Technology (KAUST), and the SDAIA-KAUST Center of
 Excellence in Data Science and Artificial Intelligence (SDAIA-KAUST AI).}
 \thanks{The authors are with the Computer, Electrical and Mathematical Sciences and Engineering, King Abdullah University of Science and Technology (KAUST), Thuwal 23955, Saudi Arabia. {\tt \{david.alvear, george.turkiyyah, shinkyu.park\}@kaust.edu.sa}}
 }

\begin{document}

\maketitle
\thispagestyle{empty}
\pagestyle{empty}

\begin{abstract}

We propose a novel framework for decision-making in cooperative grasping for two-robot object transport in constrained environments. The core of the framework is a \textit{Conditional Embedding (CE)} model consisting of two neural networks that map grasp configuration information into an embedding space. The resulting embedding vectors are then used to identify feasible grasp configurations that allow two robots to collaboratively transport an object. To ensure generalizability across diverse environments and object geometries, the neural networks are trained on a dataset comprising a range of environment maps and object shapes. We employ a supervised learning approach with negative sampling to ensure that the learned embeddings effectively distinguish between feasible and infeasible grasp configurations. Evaluation results across a wide range of environments and objects in simulations demonstrate the model's ability to reliably identify feasible grasp configurations. We further validate the framework through experiments on a physical robotic platform, confirming its practical applicability.

\end{abstract}

\section{Introduction}

\begin{figure}
  \centering
  \includegraphics[trim={0 100 0 60}, clip, width=0.43\textwidth]{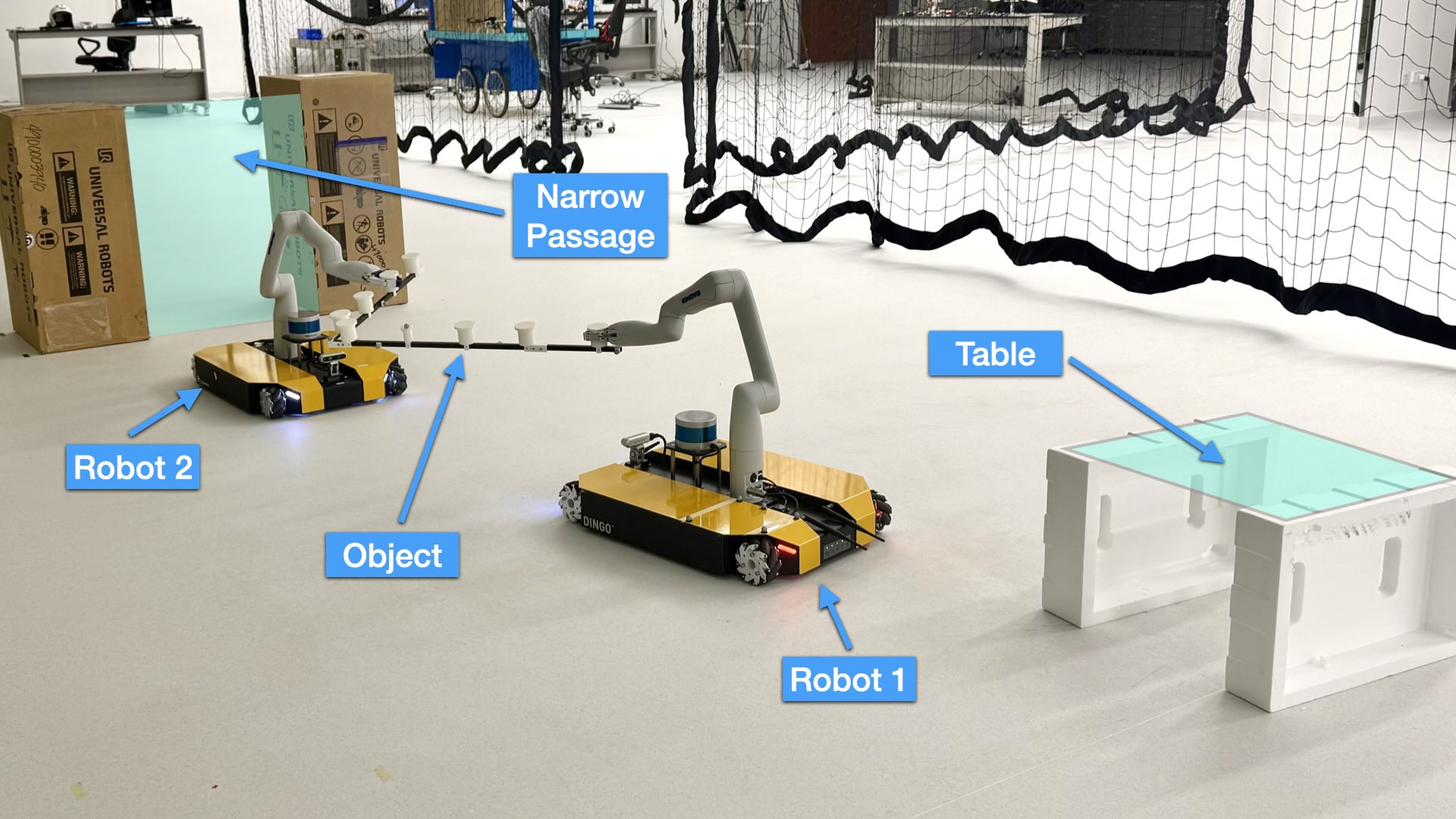}
  \caption{Object transport using two mobile manipulators. Each robot selects a grasp on the object to collaboratively move it from its initial location on the table to a designated destination, navigating through a narrow passage.}
  \label{fig:multirobot_intro}
  \vspace{-1.5em}
\end{figure}

Collective object transport is a challenging problem that involves determining the optimal grasp strategy---specifically, where and how each robot should grasp the object---to enable collaborative manipulation and transportation to a target location. This task demands precise coordination to ensure the object is securely grasped and that a feasible trajectory to the target destination exists. The complexity of identifying suitable grasp configurations increases significantly with the geometric intricacy of both the object and the surrounding environment. In this work, we propose a learning-based framework for two-robot grasp configuration selection, designed to enable reliable object transport in constrained environments (see Fig.~\ref{fig:multirobot_intro}).

Our proposed framework begins by generating a set of candidate grasp configurations, where each configuration defines a grasp point on the object for each robot and corresponding robot position required to execute the grasp. To identify viable configurations for object transport, we propose a \textit{Conditional Embedding (CE)} model. This model comprises two neural networks that project grasp configuration information into a shared embedding space, enabling the evaluation of candidate pairs through a similarity metric. The model is trained to assign high similarity scores to feasible pairs while penalizing infeasible pairs---even in the presence of class imbalance in the training data.

We validate the effectiveness of the proposed framework through extensive physics-based simulations and demonstrate its practical applicability on a physical robotic platform. While our primary contribution focuses on the design and validation of two-robot object transport, we also illustrate how the same framework can be extended to scenarios involving more than two robots by generalizing the CE model's training objective.

\subsubsection*{Related Work} \label{sec:related_work}

A wide range of studies on multi-robot object manipulation and transportation are relevant to our work. For example, Eoh \textit{et al.} \cite{eoh_multi-robot_2011} investigated push-pull strategies for transporting heavy objects using coordinated multi-robot formations. In grasp planning, Vahrenkamp \textit{et al.} \cite{vahrenkamp_planning_2010} introduced a multi-robot RRT-based planner that runs parallel bimanual inverse kinematics RRT (IK-RRT) instances, guided by inverse reachability maps to probabilistically sample feasible grasp pairs. Grasp candidates are filtered using a wrench-based threshold before computing collision-free trajectories. Similarly, Tariq \textit{et al.} \cite{tariq_grasp_2018} proposed a grasp planner that minimizes a wrench-based metric to facilitate effective load sharing in collaborative manipulation. Muthusamy \textit{et al.} \cite{muthusamy_decentralized_2014, muthusamy_task_2015, muthusamy_strictly_2023} developed a decentralized cooperative grasping framework that generates grasp candidates and selects among them using a wrench-based quality metric. Additionally, Nachum \textit{et al.} \cite{nachum_multi-agent_2019} proposed a hierarchical policy framework that separates control into two levels: a low-level policy for individual robot locomotion and a high-level policy for coordinating multi-robot manipulation. 

Existing studies on multi-robot object transport can be broadly categorized into decentralized/distributed control, optimization-based planning, and learning-based methods. In decentralized approaches, Chen \textit{et al.} \cite{chen_occlusion_2015} proposed a swarm-based approach in which robots uniformly surround an object and use local sensing to transport it toward a target, guided by the “occlusion” of the target location. Habibi \textit{et al.} \cite{habibi_centroidmr_2015} developed a distributed control strategy using local communication and centroid estimation, enabling dexterous transport via four motion controllers. Similarly, Farivarnejad \textit{et al.} \cite{farivarnejad_slidingcontrol_2016} implemented a sliding-mode controller for cooperative payload transport that relies solely on local velocity and heading measurements. While effective in decentralized coordination, these methods do not address optimal robot placement for grasping and transporting objects in constrained environments.

In optimization-based planning, Alonso-Mora \textit{et al.} \cite{mora_formation_2017} introduced a formation control framework that enables object transport in environments with both static and dynamic obstacles. Koung \textit{et al.} \cite{koung_hqp_2021} employed hierarchical quadratic programming (HQP) for coordinated object transport under rigid formation constraints. Vlantis \textit{et al.} \cite{vlantis_spacedecomp_2022} proposed a hierarchical space decomposition method to facilitate object transport in cluttered planar environments. While these approaches effectively account for environmental constraints, they typically rely on a fixed or limited set of initial robot formations and assume fixed object geometries. Consequently, they do not address the challenge of computing feasible grasp configurations for objects with diverse shapes and in environments with complex spatial constraints.

In learning-based approaches, Eoh and Park \cite{eoh_curriculum_2021} presented a deep reinforcement learning framework for multi-robot object transport, using a region-growing curriculum and a single-to-multiple-robot training scheme, where robots interact with the object via pushing in a discrete action space. Zhang \textit{et al.} \cite{zhang_descentralcon_2020} proposed a decentralized control strategy where each robot employs a Deep Q-Network (DQN) to translate sensory inputs into control commands for transporting a long rod through a narrow doorway. While effective for learning control policies, these methods typically assume fixed robot formations and do not adapt to variations in object geometry and environmental layout, limiting their applicability to more complex scenarios involving adaptive grasping.

Distinct from prior work, our study focuses on identifying feasible grasp configurations for mobile manipulators operating in constrained environments---specifically during object transport through narrow passages. To address this challenge, we propose the CE model trained via supervised learning on labeled data generated from a physics-based simulator and a negative sampling technique, 
which significantly accelerates the training. 
The model directly predicts feasible grasp configurations with high accuracy, avoiding exhaustive candidate evaluation and substantially reducing the computational overhead of trajectory planning.

\subsubsection*{Paper Organization}
In Section~\ref{section:problem_formulation}, we introduce the preliminaries and formalize the problem of the cooperative grasping for object transport. Section~\ref{section:grasping_decision_making} presents the proposed framework and introduces the CE model as a solution to the problem. In Section~\ref{sec:evaluation}, we provide simulation results to evaluate the performance of our framework and demonstrate its applicability on a physical robot platform. Section~\ref{sec:conclusions} concludes the paper.

\section{Preliminaries and Problem Description} \label{section:problem_formulation}

\begin{figure}[t]
  \centering
  \subfigure[] {
  \includegraphics[width=0.23\textwidth]{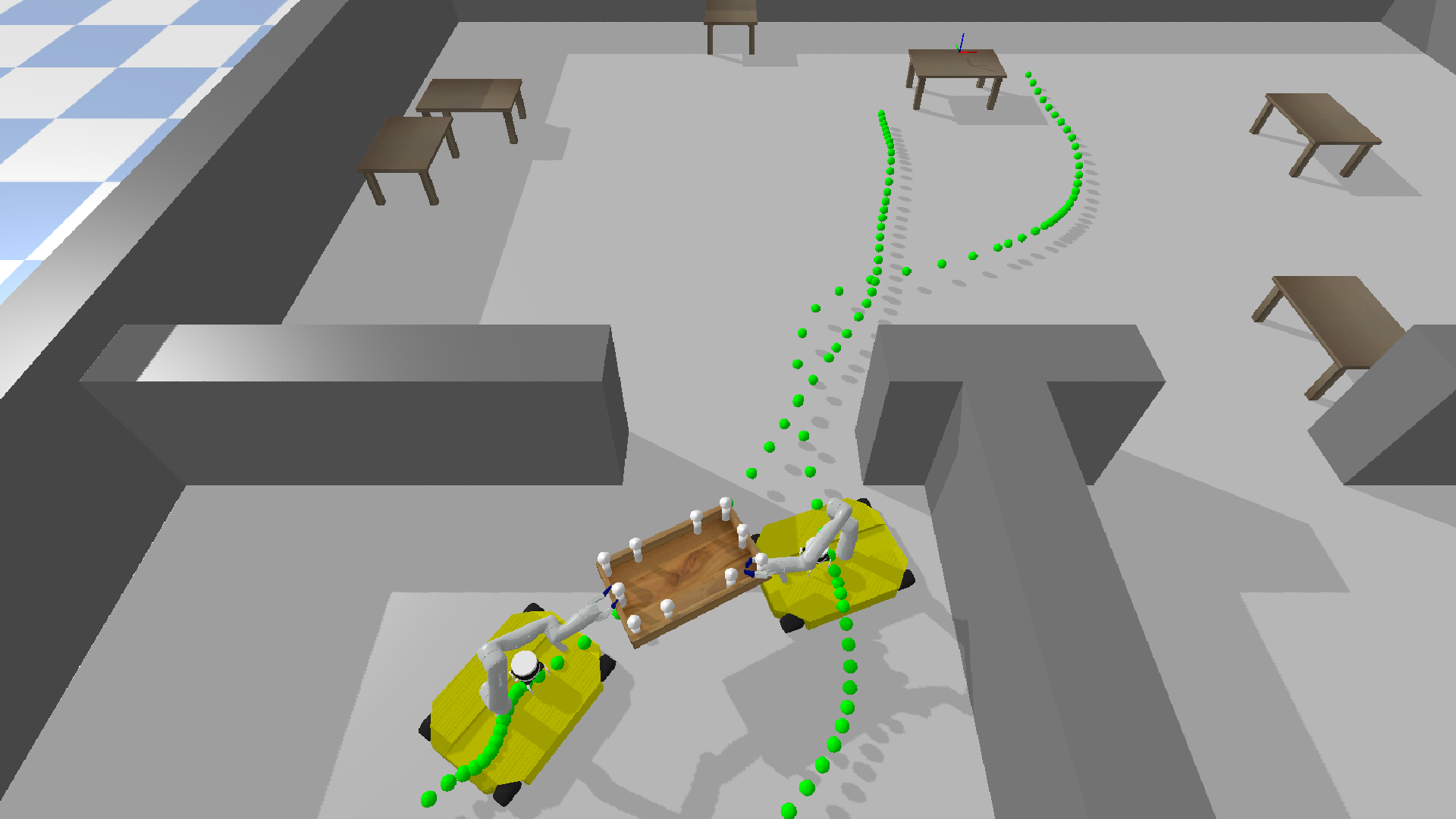}
  \label{fig:narrowpassage}
  }~~
  \subfigure[] {
  \includegraphics[width=0.14\textwidth]{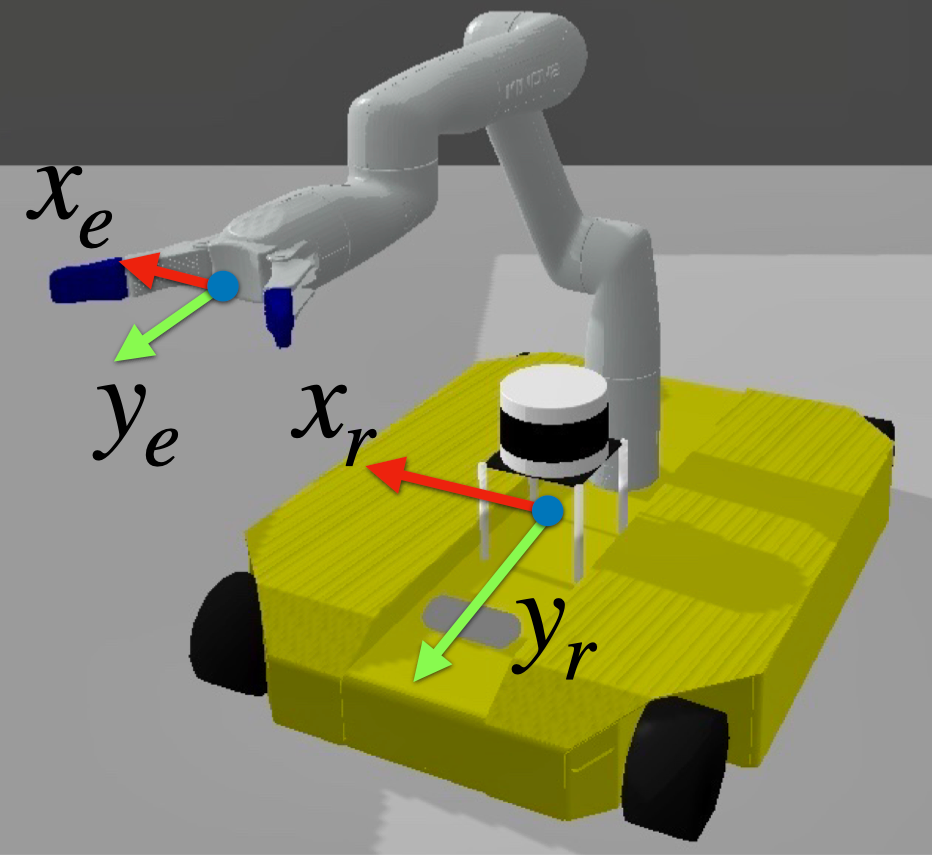}
  \label{fig:robotcoord}
  }
  \caption{(a) A collective object transport task involving two mobile manipulators navigating a narrow passage while carrying a large object that requires both robots to transport. (b) A mobile manipulator used in this work.}
  \vspace{-1.5em}
\end{figure}

The environment includes walls and randomly placed tables, which act both as obstacles and as platforms on which the object is initially positioned. It also features narrow passages (Fig.~\ref{fig:narrowpassage}), introducing additional constraints on navigation and object transport. The robot (Fig.~\ref{fig:robotcoord}) is equipped with an omnidirectional mobile base operating in a 2D plane, with position denoted by $(x_r, y_r)$. The system also includes a manipulator with a gripper, whose end-effector position is given by $(x_e, y_e, z_e)$. During transport, the gripper operates at a fixed height $z_e=0.45~m$, allowing control only in the $x$- and $y$-directions. The object is allowed to rotate only about its yaw axis.

\subsection{Definitions}

\subsubsection*{Environment Map and Object Geometry} 
Let $\mathbb{M} \subset \mathbb{R}^2$ denote the map of the environment, partitioned into the free space $\mathbb{M}_{\text{free}}$ and the obstacle region $\mathbb{M}_{\text{obs}}$. The object’s geometry is modeled as a 2D polygon from a top-down view, defined by a set of vertices $\mathbb{V} = \{v^1, \cdots, v^h\}$ that describe its boundary, where each vertex $v \in \mathbb V$ lies on the object’s perimeter.

\subsubsection*{Grasp Configurations} 
Let $\mathbb F = \{F^1, \cdots, F^n \} \subset \mathbb R^2$ denote a set of predefined grasp points distributed along the object's boundary, where each $F = (x_g,y_g) \in \mathbb F$ defines a feasible grasp point. The number of points $n$ is chosen to ensure sufficient coverage of the object's perimeter, enabling possible grasps from all directions. Given an object with $\mathbb F$ as its potential grasp points, we define the set of grasp configurations as $\mathbb G = \{G^1, \cdots, G^m\} \subset \mathbb R^2 \times \mathbb F$, where each configuration $G \in \mathbb{G}$ is a tuple $G = (x_r, y_r, x_g, y_g)$. Here, $(x_r, y_r) \in \mathbb M_{\text{free}}$ represents the collision-free position of the mobile base from which the end-effector can reach a grasp point $(x_g, y_g) \in \mathbb F$. Details on the construction of grasp configuration set are provided in Appendix~\ref{sec:grasping_candidate_sampling}.

\subsubsection*{Feasible Grasp Configurations and Trajectory Planner} Our framework draws inspiration from the Skip-Gram model in natural language processing, which captures the relationship between a \textit{center} word and its surrounding \textit{context} words. Analogously, we refer to the first robot's grasp as the \textit{center} grasp $G_{\text{center}}$ and the second robot's grasp as the \textit{context} grasp $G_\text{context}$. We define a binary metric $S(G_{\text{center}}, G_\text{context}) \in \{0, 1\}$ to indicate whether two robots can successfully transport an object to its destination using the grasp configurations $G_{\text{center}}$ and $G_\text{context}$. The metric is determined by evaluating the outcome of a trajectory planner, which computes the trajectories of two robots transporting the object, as detailed in Appendix~\ref{section:trajectory_planning}. A value of $S=1$ indicates that a feasible transportation trajectory exists, while $S=0$ denotes failure.

\subsection{Problem Description}

Given $\Phi = (\mathbb M, \mathbb V, \mathbb G )$, the objective is to determine a pair of the grasp configurations $(G_{\text{center}}, G_\text{context} )$ that enable two mobile manipulators to execute an object transport task. We propose a learning-based framework in which a policy $\pi_\theta$, conditioned on $\Phi$, is trained to predict an affinity matrix $A \in  \real^{m \times m}$, such that $A = \pi_{\theta}(\Phi)$. The affinity matrix $A$ encodes the feasibility and synergy of all possible pairs of grasp configurations for successful object transport. Once the affinity matrix $A$ is obtained, the optimal grasp configuration pair $(G_{\text{center}}^\ast, G_\text{context}^\ast )$ is selected by identifying the index pair $(i^\ast, j^\ast)$ that maximizes the affinity score: $(i^\ast, j^\ast) = \argmax_{ 1 \leq i,j \leq m} A_{ij}$, where $A_{ij}$ denotes the $i,j$-th entry of $A$, and $G_{\text{center}}^\ast = G^{i^\ast}$ and $G_{\text{context}}^\ast = G^{j^\ast}$. Alternatively, the matrix $A$ can be used to retrieve the top-$k$ grasp configuration candidates for downstream evaluation or selection.

Note that a brute-force approach, which exhaustively evaluates all possible grasp configuration pairs, requires $\frac{m!}{(m - N)!}$ evaluations to identify the optimal pair, where $N$ is the number of robots (with $N = 2$ in our case) and $m$ is the number of candidate grasp configurations. As a result, the computational complexity grows quadratically with $m$, i.e., $O(m^2)$. This highlights the significant computational burden associated with identifying feasible configuration pairs when the number of candidates is large.

\section{Cooperative Grasping Framework}
\label{section:grasping_decision_making}

\begin{figure*}[t]
  \centering
  \includegraphics[width=0.96\textwidth]{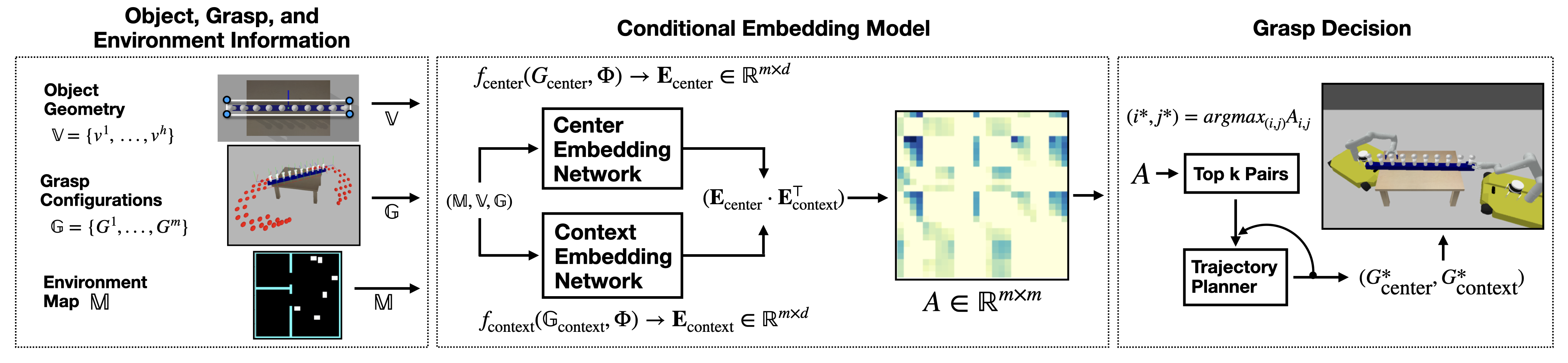}
  \caption{Grasp configuration pair selection using the CE model: Candidate grasp configurations for two mobile manipulators are generated (left block). Two embedding networks compute an affinity matrix $A$ that scores all possible configuration pairs (middle block). The top-$k$ pairs, ranked by affinity score, are selected and evaluated for feasibility using a trajectory planner (right block).}
  \label{fig:graspplanner}
  \vspace{-1.5em}
\end{figure*}
\begin{figure}
  \subfigure[]{
  \includegraphics[trim={2cm 5.5cm 1cm 5cm}, clip, width=0.49\textwidth]{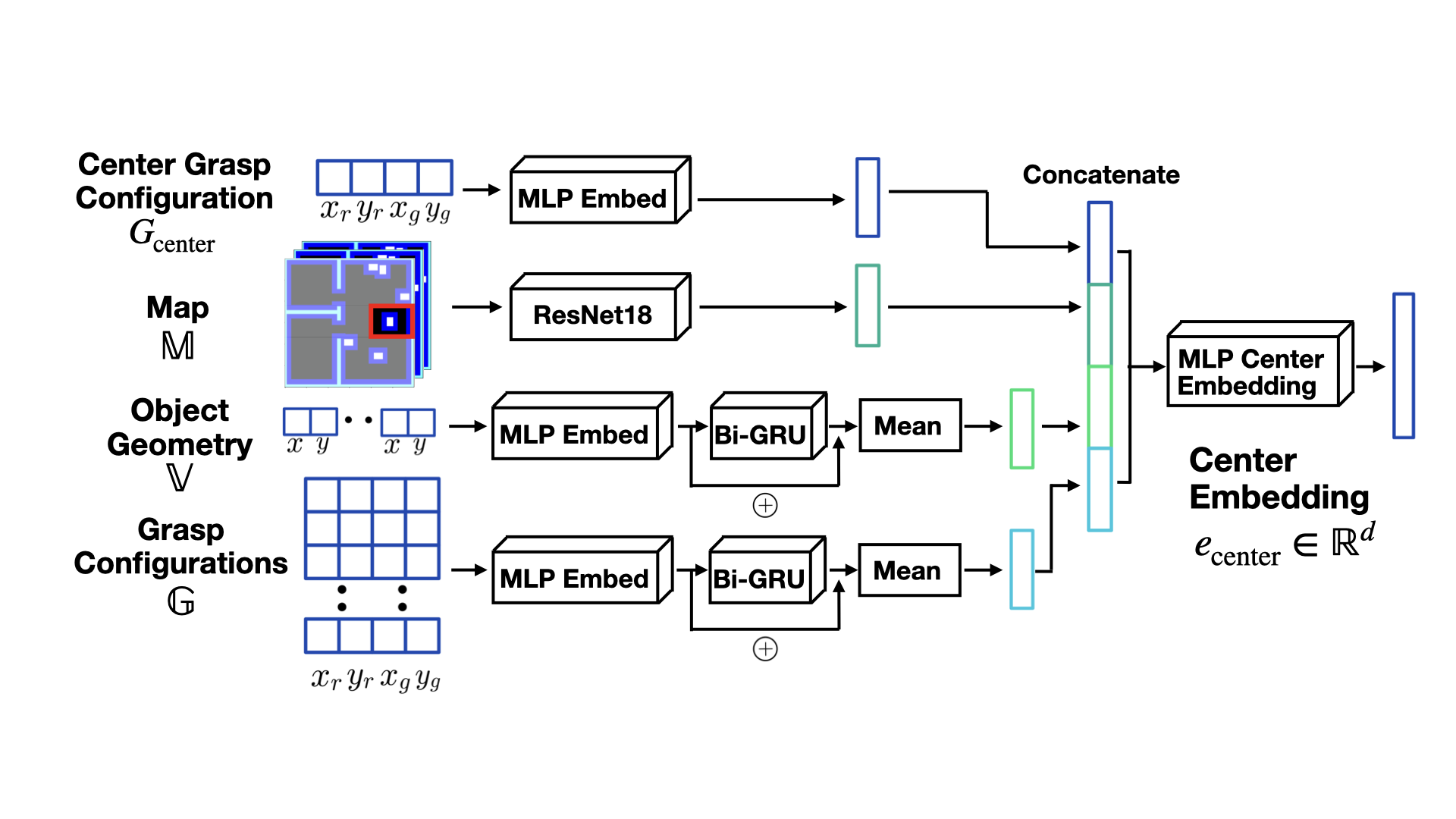}
  \label{fig:nextactcenter}
  }
  \hfill
  \subfigure[]{
  \includegraphics[trim={1cm 6cm 1cm 6.7cm}, clip, width=0.49\textwidth]{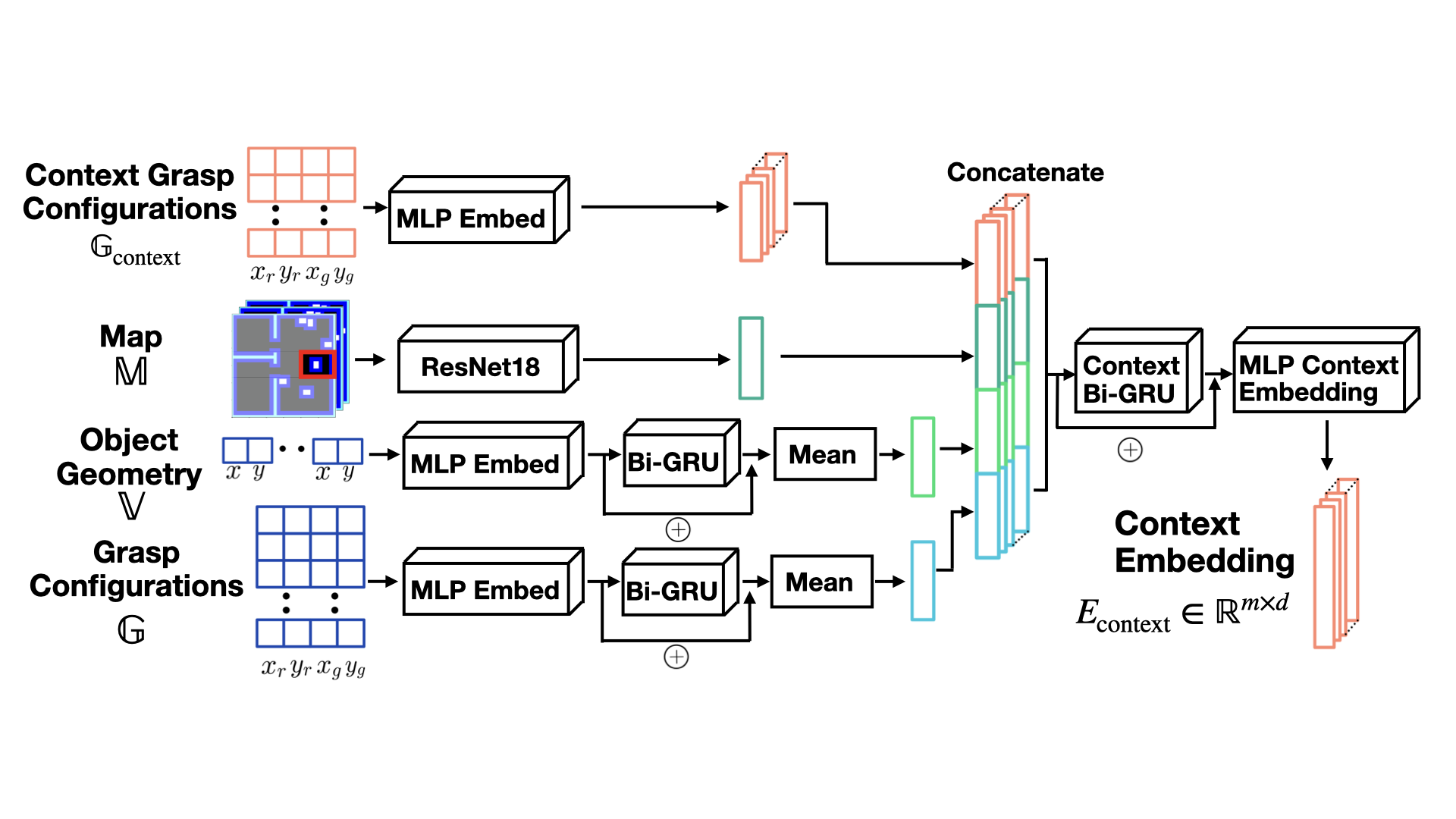}
  \label{fig:nextactcontext}
  }
  \caption{Architectures for (a) the center embedding network and (b) the context embedding network.
  }
  \label{fig:network_architecture}
  \vspace{-1.4em}
\end{figure}

In the proposed framework, the trained CE model takes the input state $\Phi$ and generates $k$ candidate grasp configuration pairs. Each pair is then evaluated by the trajectory planner to determine whether a collision-free path for object transport exists. The overall pipeline for identifying feasible grasp configurations using the CE model is illustrated in Fig.~\ref{fig:graspplanner}.

\subsection{Conditional Embedding (CE) Model}

The CE model consists of two neural networks: the \textit{center embedding} network and \textit{context embedding} network (see Figs.~\ref{fig:graspplanner} and \ref{fig:network_architecture}). Each network maps the grasp configuration information into an embedding space $\real^d$, producing a pair of embeddings. Here, $d$ is a configurable parameter that defines the dimensionality of the embedding space. The model is trained such that the embeddings corresponding to a grasp configuration pair $(G_{\text{center}}, G_\text{context})$ are positively correlated if the pair is feasible for object transport, i.e., $S(G_{\text{center}}, G_\text{context}) = 1$. 

\subsubsection*{Center Embedding Network $f_{\text{center}}$}
The center embedding vector $\mathbf{e}_{\text{center}} \in \mathbb{R}^d$ (with unit length) for $G_{\text{center}} \in \mathbb G$ is generated by a function $f_{\text{center}}$: $\mathbf{e}_{\text{center}} = f_{\text{center}}(G_{\text{center}}, \Phi)$. The function $f_{\text{center}}$ is implemented as a neural network, whose architecture is illustrated in Fig.~\ref{fig:nextactcenter}. Elements of the object geometry set $\mathbb{V}$ and the grasp configuration set $\mathbb{G}$ are first passed through fully connected multi-layer perceptron (MLP) heads in parallel, followed by bidirectional gated recurrent units (Bi-GRU) to capture sequential dependencies, and subsequently aggregated using mean operators. Their outputs are concatenated with feature vectors of the center grasp configuration $G_{\text{center}}$ and the environment map $\mathbb M$, which are processed using another MLP head and a ResNet18 encoder, respectively. The resulting concatenated feature vector is then further processed and normalized to produce the unit-length center embedding vector $\mathbf e_{\text{center}}$.

\subsubsection*{Context Embedding Network $f_{\text{context}}$}
This network processes a subset of context grasp configurations $\mathbb{G}_{\text{context}} \subseteq \mathbb{G}$ and outputs their embeddings as a matrix $\mathbf{E}_{\text{context}} = f_{\text{context}}(\mathbb{G}_{\text{context}}, \Phi)$, where each row vector $\mathbf{e}_{\text{context}} \in \mathbb{R}^d$ represents the embedding of a grasp configuration $G_{\text{context}} \in \mathbb{G}_{\text{context}}$. The function $f_{\text{context}}$ is implemented as a neural network, with its architecture shown in Fig.~\ref{fig:nextactcontext}. The environment map $\mathbb{M}$, object geometry set $\mathbb{V}$, and full grasp configuration set $\mathbb{G}$ are processed in the same manner as in the center embedding network. Each element of $\mathbb{G}_{\text{context}}$ is first passed through a shared MLP head in parallel, producing individual feature vectors. These are concatenated with the feature representations of $\mathbb{M}$, $\mathbb{V}$, and $\mathbb{G}$ to form a set of larger feature vectors, which are then processed sequentially by a Bi-GRU, followed by an MLP and normalization layer. The final output is the matrix $\mathbf{E}_{\text{context}}$, whose rows are the unit-length context embedding vectors for $\mathbb G_{\text{context}}$.

The similarity between the embeddings $\mathbf{e}_{\text{center}}$ and $\mathbf{e}_{\text{context}}$ is used to estimate the conditional probability that a $G_{\text{context}}$ is compatible with $G_{\text{center}}$:
\begin{align}
    \label{eq:conditionalembedding}
    \mathbb P (G_{\text{context}} \,|\, G_{\text{center}}) = \frac{\exp (\mathbf{e}_{\text{context}} \cdot \mathbf{e}_{\text{center}} )}{\sum_{j=1}^{m} \exp (\mathbf{e}_{\text{context}}^{j} \cdot \mathbf{e}_{\text{center}} )},
\end{align}
where $\mathbf{e}_{\text{context}}^{j}$ is the context embedding corresponding to $G^j \in \mathbb G$ and $m = |\mathbb G|$ is the total number of candidate grasp configurations.

\subsection{Model Training}
All model parameters are optimized during training, except for the weights of the ResNet18 backbone; only the final fully-connected head layers are fine-tuned, while the preceding convolutional blocks remain frozen. Layer normalization is applied after every hidden linear layer in the MLP-Embed and in the final MLPs that generate the center and context embeddings, providing regularization and stabilizing optimization. The CE model is trained by maximizing the following log-likelihood:
\begin{align}
    \label{eq:likelihoodembedding}
    \textstyle \sum_{G_{\text{center}} \in \mathbb G} \sum_{G_{\text{context}} \in \mathcal {DC}(G_{\text{center}})} \mathbb \log \, \mathbb P(G_{\text{context}} \,|\, G_{\text{center}}),
\end{align}
where the conditional probability $\mathbb P(G_{\text{context}} \,|\, G_{\text{center}})$ is given by the softmax formulation in \eqref{eq:conditionalembedding}. The \textit{Dynamic Context (DC)} set $\mathcal {DC}(G_{\text{center}}) \subseteq \mathbb G$ includes all context grasp configurations $G_\text{context}$ that satisfy the feasibility condition $S(G_{\text{center}}, G_\text{context})=1$.\footnote{The size of $\mathcal {DC}(G_{\text{center}})$ varies depending on $G_{\text{center}}$, and may contain many, few, or no feasible pairings.}

However, directly computing the softmax denominator in \eqref{eq:conditionalembedding} becomes computationally expensive when the number of candidate configurations $m$ is large. To mitigate this, we adopt a negative sampling-based training strategy inspired by the Word2Vec framework \cite{zhang_dive_2023, murphy_kevin_p_dimensionality_2022, mikolov_efficient_2013}, which replaces the softmax training objective with a set of binary classification problems. 
Specifically, we introduce a binary random variable $D \in \{0, 1\}$ for each grasp pair $(G_{\text{center}}, G_{\text{context}})$, where $D=1$ if $G_{\text{context}} \in \mathcal {DC} (G_{\text{center}})$ and $D=0$ otherwise. The conditional probability of this event is modeled using a sigmoid function applied to the dot product of the corresponding embedding vectors: \vspace{-.2em}
\begin{align}
    \label{eq:D1Event}
    \mathbb P(D\!=\!1 | G_{\text{center}}, G_{\text{context}}) \!=\! \frac{1}{1 \!+\! \exp(-\mathbf{e}_{\text{context}} \!\cdot\! \mathbf{e}_{\text{center}}/\tau)},
\end{align}
where $\tau$ is a temperature parameter that controls the sharpness of the sigmoid.

Using this expression, the conditional probability in \eqref{eq:conditionalembedding} is replaced by: \vspace{-.5em}
\begin{multline}    \label{eq:approxconditionalprobability}
    \mathbb P(D=1 \mid G_{\text{center}}, G_\text{context}) \\
    \textstyle\times \frac{1}{|\mathcal{N}(G_{\text{center}})|} \prod_{G \in \mathcal{N}(G_{\text{center}})} 
    \mathbb P (D=0 \mid G_{\text{center}}, G),
\end{multline}
where $\mathcal{N}(G_{\text{center}})$ is a set of negative samples---grasp configurations uniformly drawn from $\mathbb G \setminus \mathcal{DC}(G_{\text{center}})$. For each $G_{\text{center}}$, the number of negative samples is chosen such that the size of the union $\mathcal DC(G_{\text{center}}) \cup \mathcal N(G_{\text{center}})$ remains constant across all center grasp configurations.\footnote{This reformulation decouples the complexity of evaluating \eqref{eq:conditionalembedding} from the total number of grasp configurations $m = |\mathbb{G}|$, reducing it to scale with the number of negative samples, which can be chosen to be significantly smaller than $m$.} 

\subsection{Affinity Matrix Computation and Grasp Selection}
Once trained, the two neural networks are used to generate embeddings for all candidate grasp configurations in $\mathbb{G}$. The resulting embeddings are organized into two matrices: the center embedding matrix $\mathbf{E}_{\text{center}} \in \mathbb{R}^{m \times d}$ and the context embedding matrix $\mathbf{E}_{\text{context}} \in \mathbb{R}^{m \times d}$. An affinity matrix is then computed as $A = \mathbf{E}_{\text{center}} \cdot \mathbf{E}_{\text{context}}^\top$, where each entry reflects the similarity between a center-context grasp pair. The top-$k$ entries of $A$---those with the highest similarity scores—are selected as the most promising grasp configuration pairs for collective object transport. These selected pairs correspond to those that maximize the probability in \eqref{eq:D1Event}.

\subsection{Extension to Multi-Robot Object Transport}

Recall that the CE model generates embeddings for all grasp configurations in $\mathbb G$, resulting in two sets: the center embeddings $(\mathbf e_{\text{center}}^{1}, \cdots, \mathbf e_{\text{center}}^{m})$ and context embeddings $(\mathbf e_{\text{context}}^{1}, \cdots, \mathbf e_{\text{context}}^{m})$. The model can be trained to maximize the following objective, extended to the $N$-robot setting:
\begin{align*}
    \textstyle \sum_{G_{\text{center}} \in \mathbb G} \sum_{P_N \in \mathcal {DC} (G_{\text{center}})} \log \mathbb P( P_N \,|\, G_{\text{center}}).
\end{align*}

The \textit{DC} set $\mathcal {DC} (G_{\text{center}}) \subset \mathbb G^{N-1}$ contains grasp configuration tuples $P_N = (G_{\text{context}}^{i_2}, \cdots, G_{\text{context}}^{i_N})$ for the remaining $N-1$ robots.
A combination $(G_{\text{center}}, G_{\text{context}}^{i_2}, \cdots, G_{\text{context}}^{i_N})$ is a feasible grasp configurations for $N$ robots if and only if $P_N \in \mathcal {DC} (G_{\text{center}})$. The conditional probability can be defined using a softmax function as:
\begin{align*}
    &\mathbb P( P_N \,|\, G_{\text{center}}) \nonumber \\
    &= \frac{\exp(\frac{1}{N-1} (\mathbf e_{\text{context}}^{i_2} + \cdots + \mathbf e_{\text{context}}^{i_N}) \cdot \mathbf e_{\text{center}} )}{\sum_{P_N' \in \mathbb G^{N-1}} \exp( \frac{1}{N-1} (\mathbf e_{\text{context}}^{i_2'} + \cdots + \mathbf e_{\text{context}}^{i_N'}) \cdot \mathbf e_{\text{center}})},
\end{align*}
where $(\mathbf e_{\text{context}}^{i_2'}, \cdots,  \mathbf e_{\text{context}}^{i_N'})$ are context embedding vectors corresponding to the grasp configuration tuple $P_N' = (G_{\text{context}}^{i_2'}, \cdots, G_{\text{context}}^{i_N'})$.

To facilitate grasp selection across $N$ robots, we define an affinity tensor $\mathcal{A}: \mathbb G \times \mathbb G^{N-1} \to \mathbb R$ as:
\begin{align*}
    \textstyle \mathcal{A} (G_{\text{center}}, P_N) = \textstyle \frac{1}{N-1} (\mathbf e_{\text{context}}^{i_2} + \cdots + \mathbf e_{\text{context}}^{i_N}) \cdot \mathbf e_{\text{center}}.
\end{align*}
Since $\mathcal{A}$ is computed as the dot product between the center embedding and the average of the context embeddings, it quantifies the collective compatibility of the grasp configurations. 
The optimal grasp configurations are then obtained by selecting the tuple $(G_{\text{center}}^\ast, P_N^\ast) = \argmax_{(G_{\text{center}}, P_N)} \mathcal{A} (G_{\text{center}}, P_N)$.

\section{Evaluation} \label{sec:evaluation}

\subsection{Simulation Setup and Model Training} \label{section:experiment_setup}

\begin{figure}[t]
    \centering
    \subfigure[Scenes]{\includegraphics[width=0.13\textwidth]{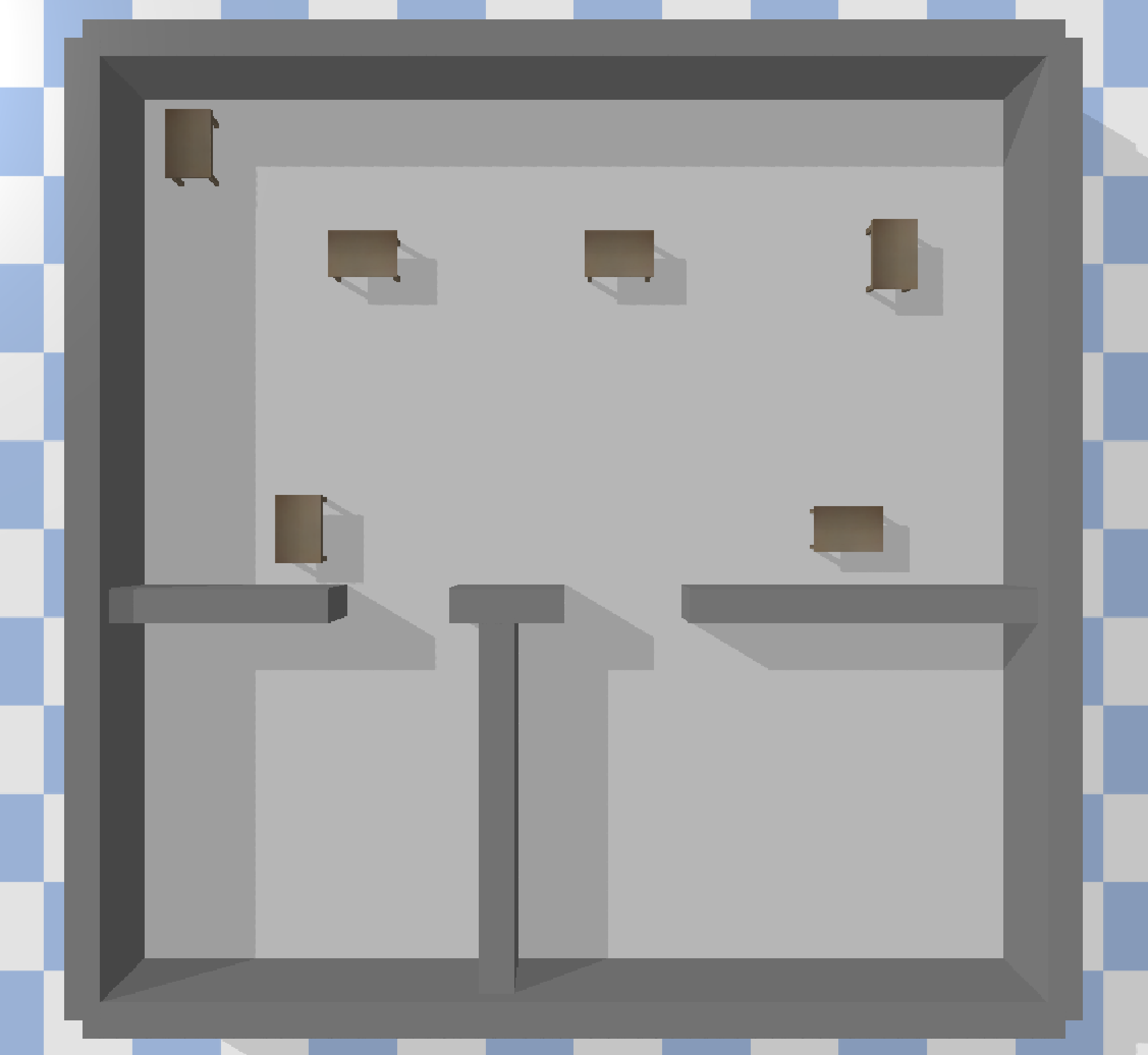}
    \includegraphics[width=0.13\textwidth]{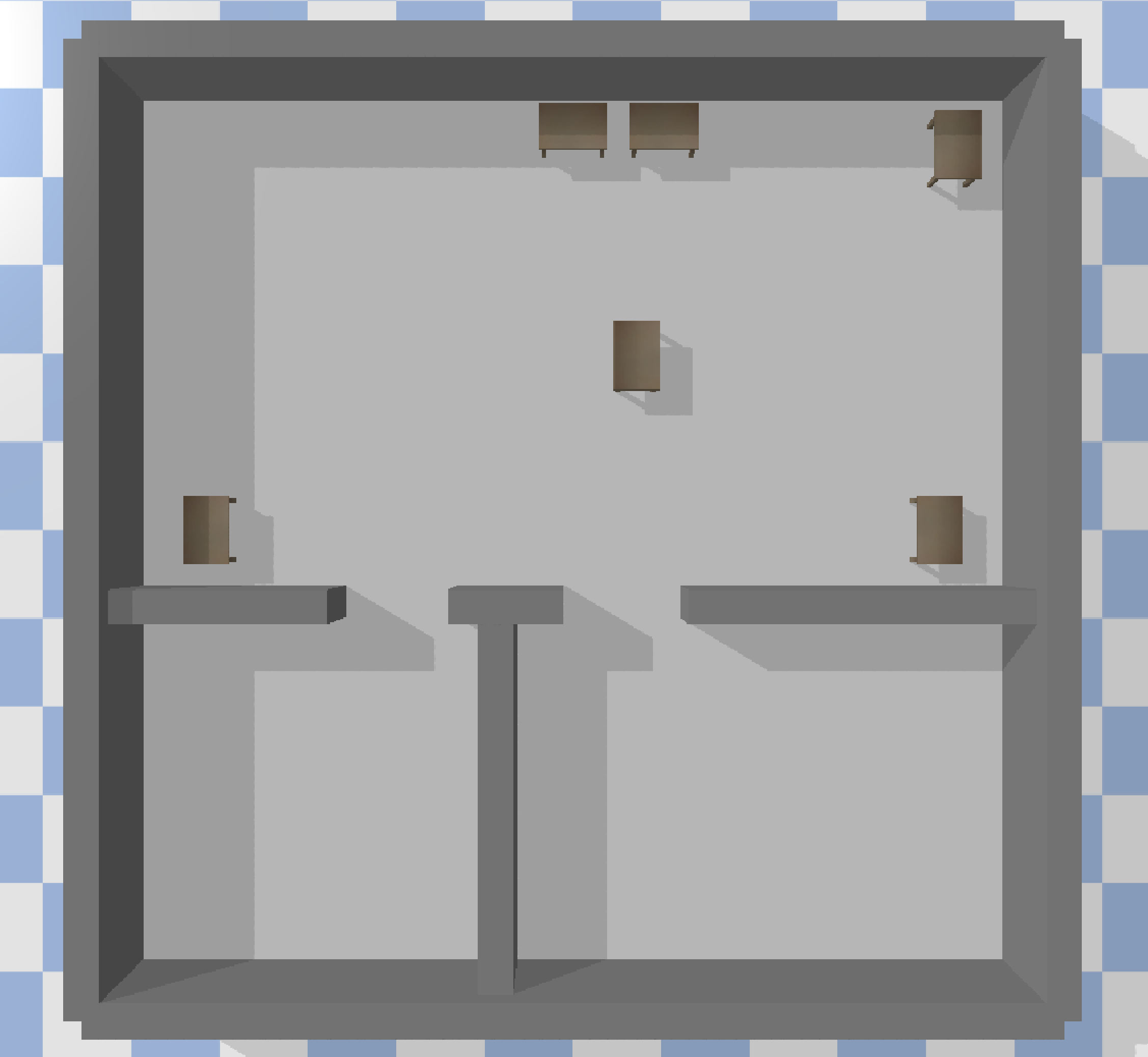}
    \label{fig:tables}}
  \subfigure[Training Objects]{%
    \begin{minipage}[b]{0.13\textwidth}
      \centering
      \includegraphics[trim={12cm 7cm 15cm 13cm}, clip,
                       height=0.46\linewidth]{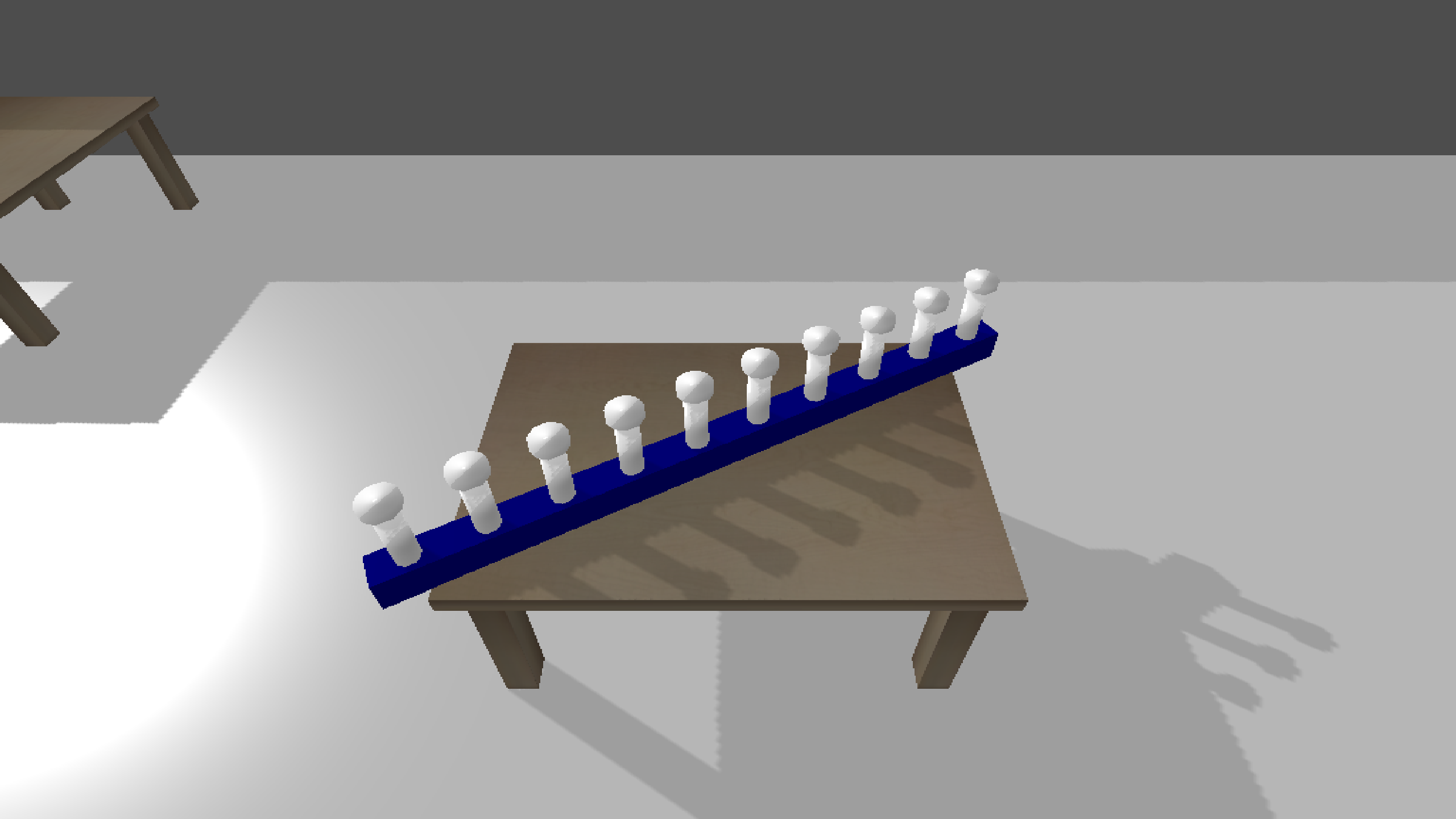}\\[1pt]
      \includegraphics[trim={12cm 7cm 15cm 13cm}, clip,
                       height=0.46\linewidth]{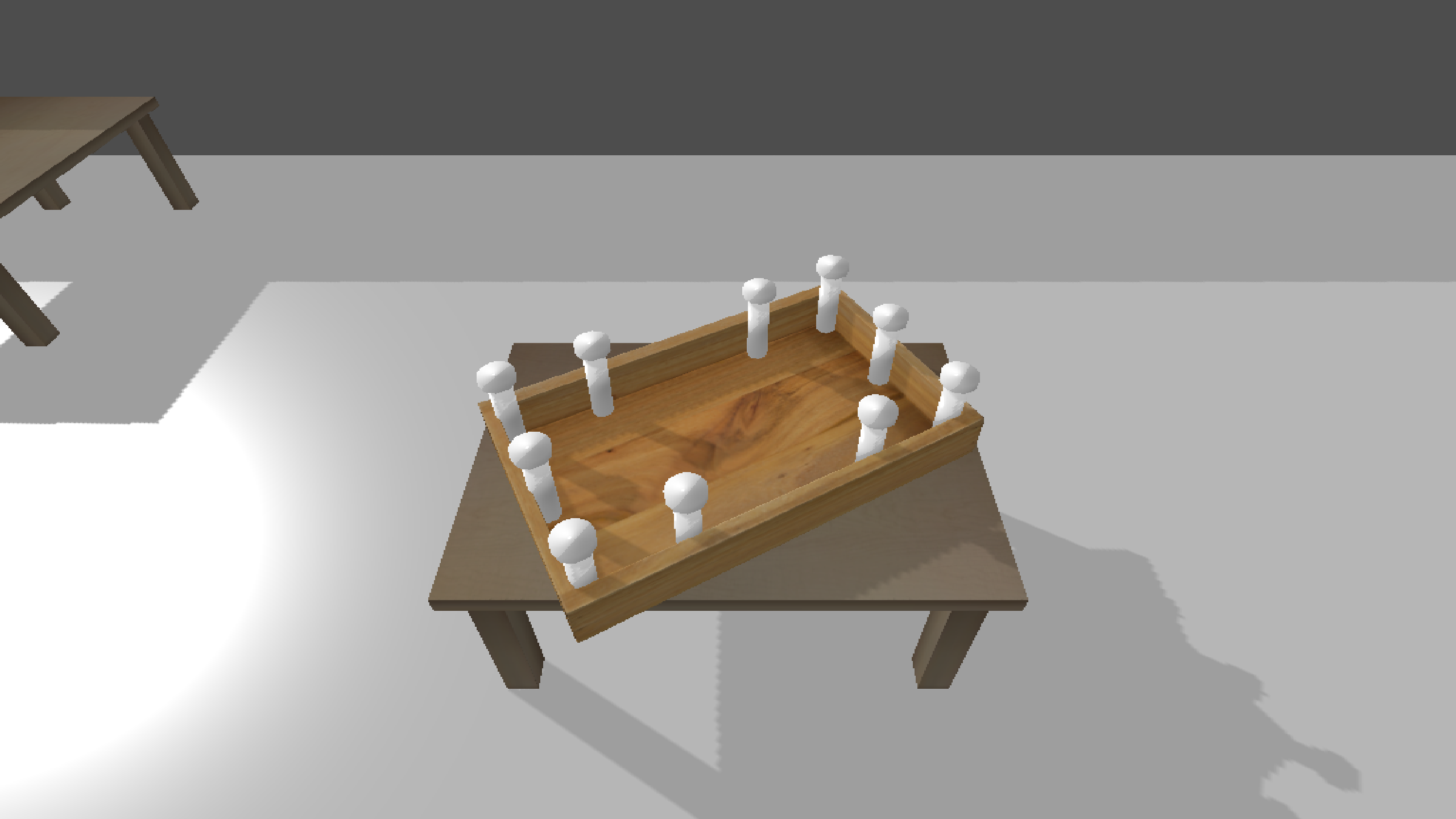}
    \end{minipage}
    \label{fig:objects}}
  \caption{Environments for collective object transport, featuring two table arrangements for object placement as illustrated in (a). The objects shown in (b) are used to generate the dataset for training the CE model, each with ten predefined grasp points for both the long bar and the rectangular object.}
  \vspace{-1.0em}
\end{figure}

For model training, we employ two distinct table setups, as shown in Fig.~\ref{fig:tables}, and two object types illustrated in Fig.~\ref{fig:objects}. The first object is a long bar with white cylindrical grasping elements, while the second is a rectangular object with similar cylindrical elements distributed around its perimeter. Each cylindrical element serves a stable grasp point and corresponds to an element in the set $\mathbb F$. 

The training dataset comprises $576$ samples, generated by placing each object on one of six tables at 24 orientations uniformly sampled over the range $[-\pi, \pi)$. Each object is initially positioned on a table, and its destination is set as the center of the nearest unoccupied room on the opposite side of the corridor. After filtering out grasp configuration candidates that are infeasible for object grasping due to collisions with obstacles, the size of the grasp configuration set $\mathbb G$ typically ranges from $20$ to $150$.

\begin{figure}[t]
  \centering
  \subfigure[]{
  \includegraphics[trim={0cm 20cm 50cm 0cm}, clip, width=0.19\textwidth]{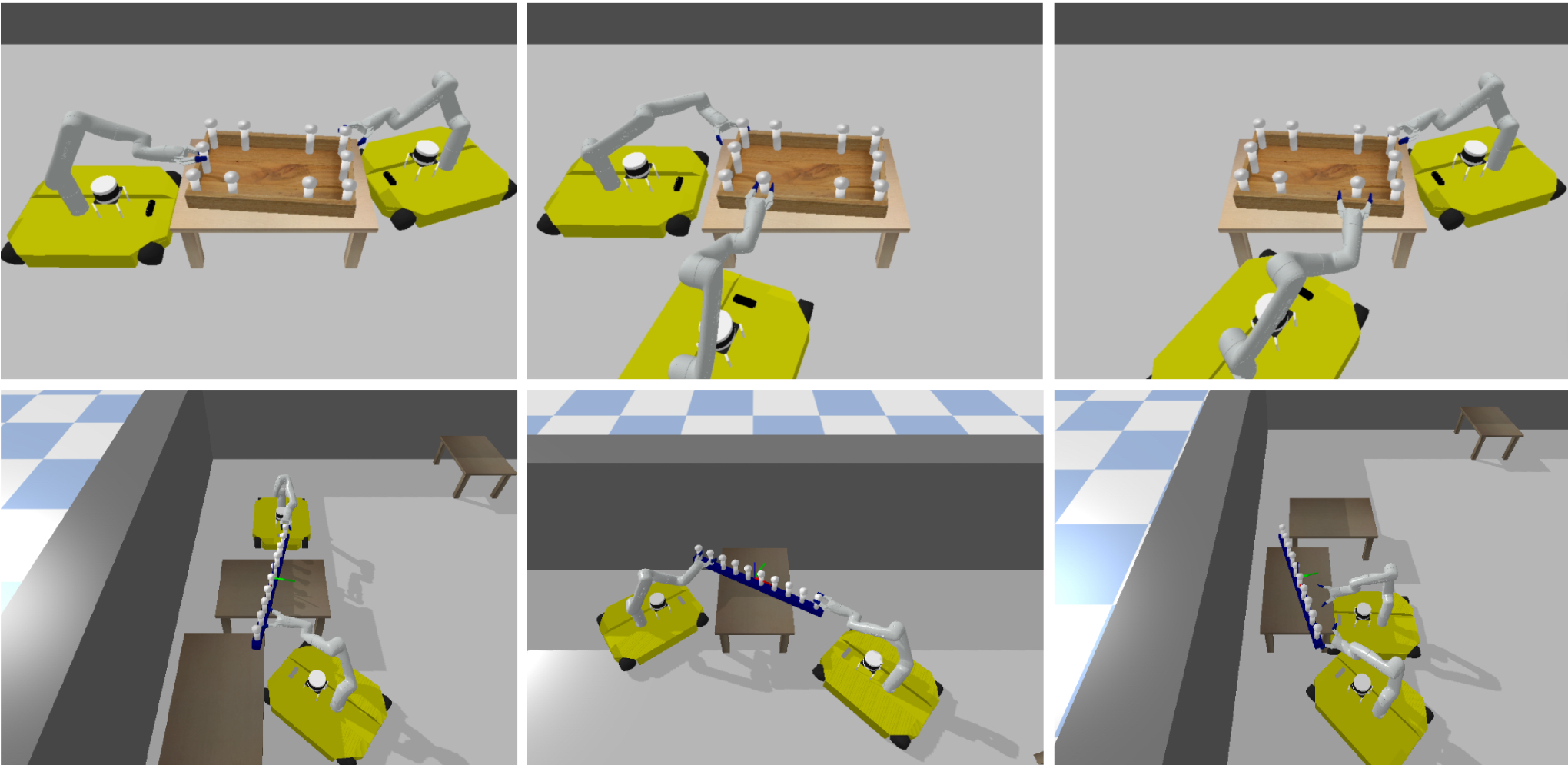}
  }
  \subfigure[]{
  \includegraphics[trim={25cm 20cm 25cm 0cm}, clip, width=0.19\textwidth]{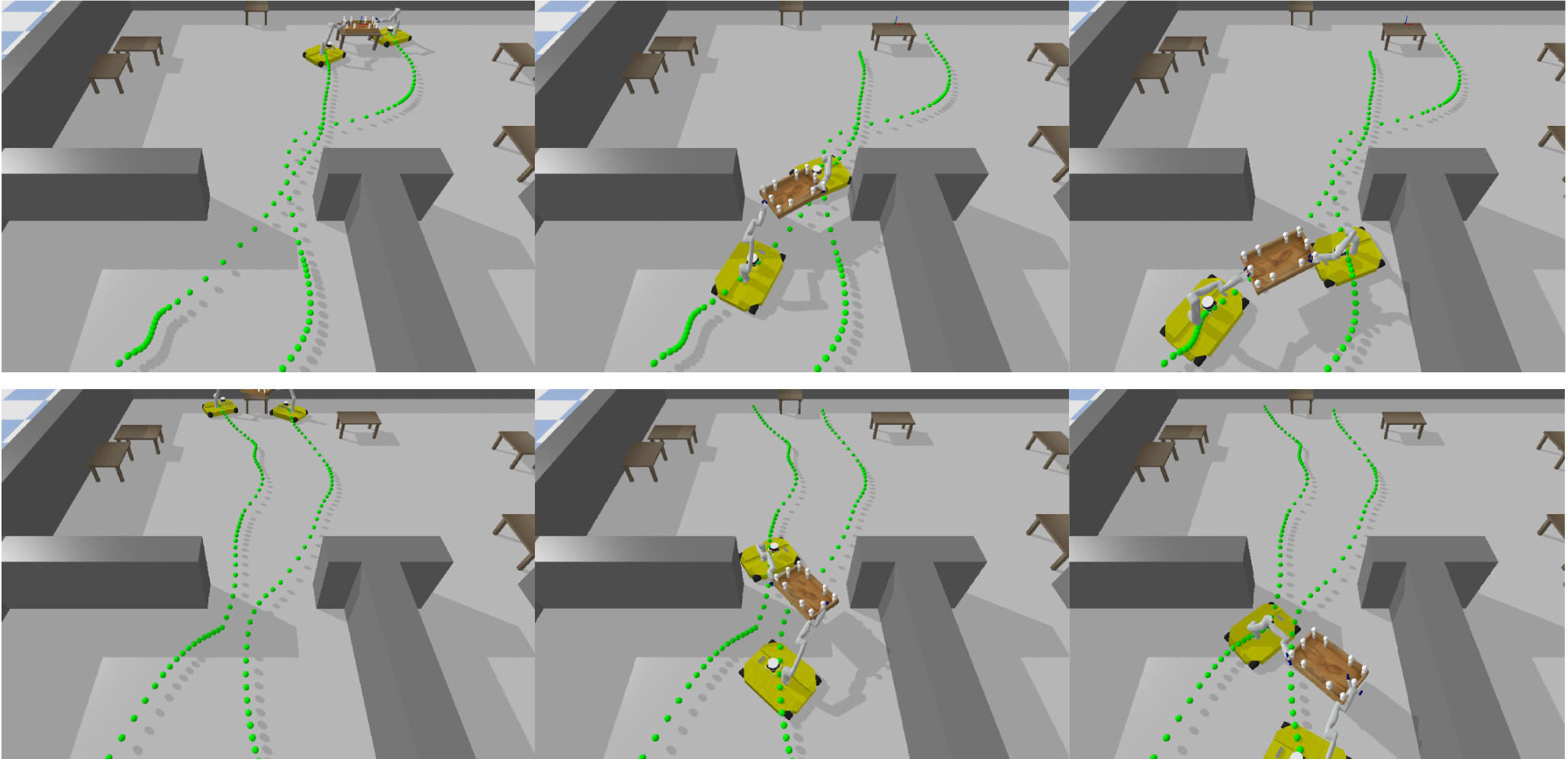}
  }
  \caption{(a) Validation of grasp configuration pairs for collective object transport in PyBullet. (b) Trajectory generation for a selected grasp configuration pair using the trajectory planner.}
  \label{fig:validation}
  \vspace{0em}
\end{figure}

For each sample, all feasible pairs of grasp configurations are identified based on their ability to successfully complete the object transport task, as determined by the trajectory planner (detailed in Appendix~\ref{section:trajectory_planning}) and illustrated in Fig.~\ref{fig:validation}. The planner is implemented using Cyipopt \cite{ipopt2006} and PyBullet \cite{coumans_pybullet_2016}. Dataset generation was performed on a high-performance computing cluster with $1000$ Intel® Xeon® Gold 6148 CPUs, with each feasibility check taking under 30 seconds.

The learning objective in \eqref{eq:likelihoodembedding} is implemented as follows:
\begin{multline}
    \label{eq:weighted_loss}
    \textstyle \sum_{G_{\text{center}} \in \mathbb{G}} \big( \sum_{G_{\text{context}} \in \mathcal{DC}(G_{\text{center}})}  \log p (G_{\text{center}}, G_{\text{context}}) \\ + \textstyle  \frac{|\mathcal{DC}(G_{\text{center}})|}{|\mathcal{N}(G_{\text{center}})|} \sum_{G \in \mathcal{N}(G_{\text{center}})} \log(1 - p (G_{\text{center}}, G)) \big),
\end{multline}
where $p(G_{\text{center}}, G) = \mathbb{P}(D=1 \mid G_{\text{center}}, G)$ defined as in \eqref{eq:D1Event}.
The negative set $\mathcal {N}(G_{\text{center}})$ is constructed to ensure that the union $\mathcal {DC}(G_{\text{center}}) \cup \mathcal {N}(G_{\text{center}})$ has a fixed size $K$ across all $G_{\text{center}}$ in $\mathbb G$. Specifically,  $K = \alpha \max_{G_{\text{center}} \in \mathbb{G}} \lvert \mathcal {DC}(G_{\text{center}})\rvert$, where $\alpha$ is a hyperparameter that controls the ratio of feasible to infeasible samples and is tuned based on validation performance. 

The full dataset is divided into $70\%$ for training, $20\%$ for validation, and $10\%$ for testing.  Hyperparameters, summarized in Table~\ref{tab:hyperparameters}, are tuned using Bayesian optimization.  The model is trained using the AdamW optimizer for $83$ epochs.

\begin{table}[t]
\setlength{\tabcolsep}{4pt}
\footnotesize
    \centering
    \caption{Hyperparameters for Training}
    \begin{tabular}{c|c||c|c}
        \toprule
        \textbf{Hyperparameters} & \textbf{Value} & \textbf{Hyperparameters} & \textbf{Value} \\
        \hline
        Learning rate & $2.61\times 10^{-4}$ & Weight decay & $1\times 10^{-4}$ \\
        Batch size & 37 & Embed size ($d$) & 44 \\
        Epochs & 83 & Temperature ($\tau$) & $6.15\times 10^{-2}$ \\
         LR‐schedule factor & 0.3741 & Scaling factor ($\alpha$) & 1.10 \\
        Schedule patience & 3 epochs & Optimizer &  AdamW\\
        \bottomrule
    \end{tabular}
    \label{tab:hyperparameters}
    \vspace{-1.0em}
\end{table}

\subsection{Evaluation Results}
Using the test dataset, we evaluate the CE model as a recommendation system by analyzing its top-$k$ performance---ranking grasp configurations based on their scores in the affinity matrix derived from the model’s embedding representations. Table~\ref{tab:topkmetrics} reports the success rate of identifying at least one feasible grasp configuration pair among the top-1, top-3, and top-5 ranked candidates. For reference, we also include a random selection policy to highlight the challenge of finding feasible pairs through random sampling. The model consistently achieves high success rates---surpassing $83\%$ across all cases and reaching up to $99.22\%$ for the top-5 candidates---compared to $41.11\%$ achieved by random sampling.

We also evaluate the CE model’s ability to distinguish between feasible and infeasible grasp configuration pairs using \eqref{eq:D1Event}. Given the class imbalance---where infeasible pairs outnumber feasible ones---a threshold of $0.64$ is applied to the output of \eqref{eq:D1Event}, selected by maximizing the F1-score. Table~\ref{tab:metrics} presents the evaluation results across various performance metrics. The reported \textit{accuracy} of $90.56\%$ indicates the proportion of correctly classified grasp configuration pairs. \textit{Precision} measures the proportion of predicted feasible pairs that are actually feasible, while \textit{recall} reflects the model’s ability to identify all true feasible pairs. The high recall indicates that the model rarely misses feasible grasp pairs, and the high precision suggests a low false positive rate. The high \textit{F1-score}---the harmonic mean of precision and recall---confirms balanced performance. Finally, the \textit{AUC-ROC} score indicates the model assigns higher scores to feasible pairs than infeasible ones $96.02\%$ of the time, confirming its strong discriminative capability.

\subsection{Generalization to Novel Environments}

\begin{figure}[t]
  \centering
  \subfigure[]{
      \includegraphics[width=0.28\textwidth, height=.09\textwidth]{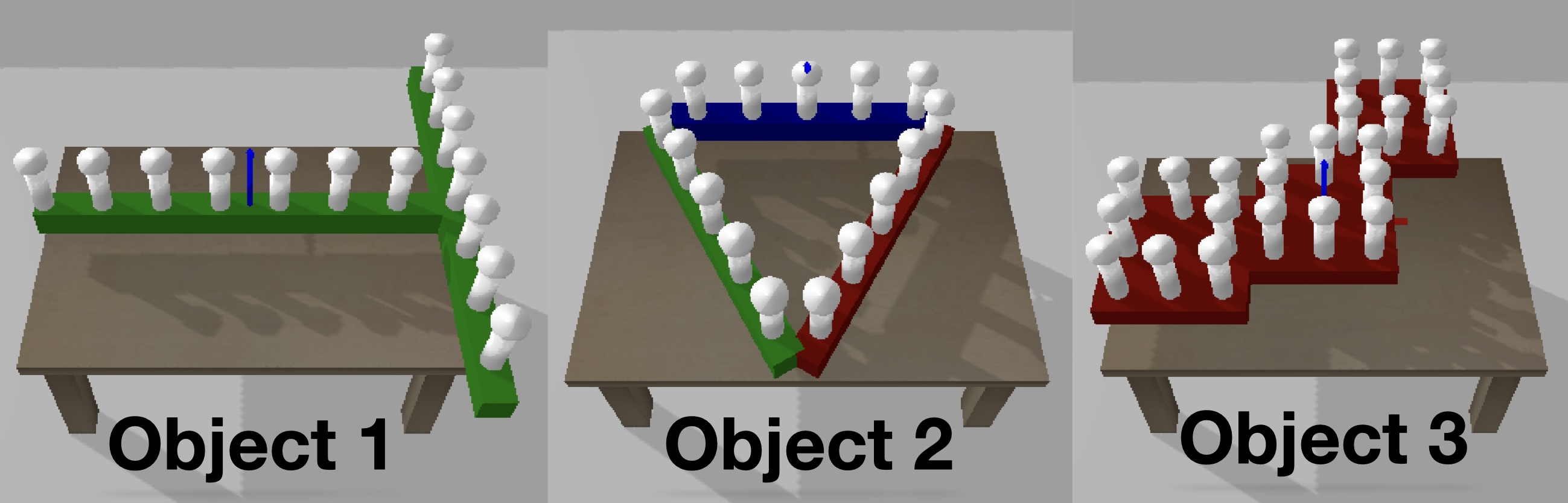}
      \label{fig:evaluationassets_a}
  }
  \subfigure[]{
      \includegraphics[width=0.14\textwidth, height=.09\textwidth]{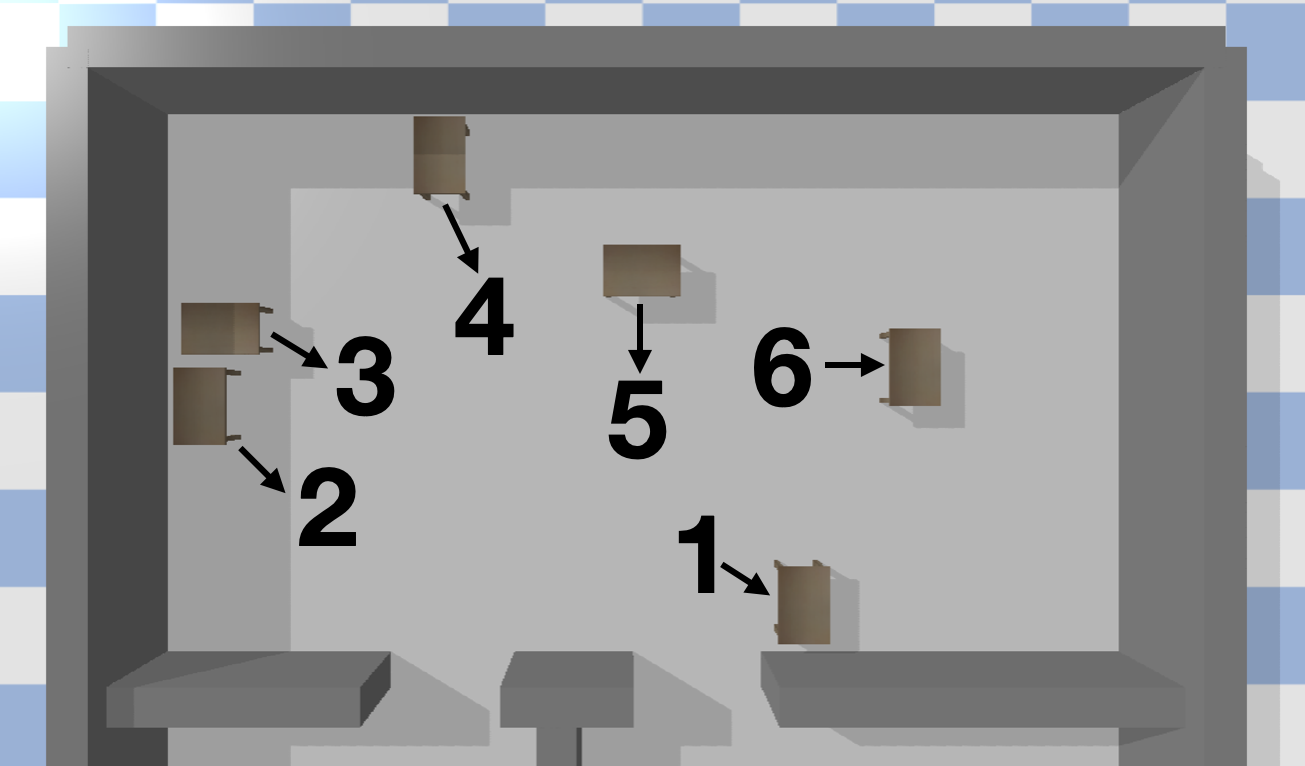}
      \label{fig:evaluationassets_b}
  }
  \caption{(a) Three additional objects with diverse geometries used for generalization testing: Object 1, a T-shaped panel with 14 grasp points; Object 2, a triangular object with 15 grasp points; Object 3, an asymmetric panel with 24 grasp points. (b) Evaluation environment, with only the table-placed region shown.}
  \label{fig:evaluationassets}
  \vspace{0em}
\end{figure}

Three new objects---each with a distinct shape and multiple grasp points, as illustrated in Fig.~\ref{fig:evaluationassets_a}---are specifically designed to assess the generalization capability of the CE model. The evaluation environment consists of six tables arranged as shown in Fig.~\ref{fig:evaluationassets_b}. Each object is placed on one of the tables at ten uniformly spaced orientations over the range $[-\pi, \pi)$, resulting in $180$ distinct evaluation scenarios. The number of candidate grasp configurations per scenario ranges from 80 to 450.

\begin{table}[t]
  \centering
    \caption{Top-$k$ Evaluation Results}
    \begin{tabular}{c|c|c|c}
    \toprule
    \textbf{Top-$k$} & \textbf{Top-$1$} & \textbf{Top-$3$} & \textbf{Top-$5$} \\
    \textbf{Method} & Success \% & Success \% &  Success \% \\
    \hline
    CE Model & 83.05 & 96.39 & 99.22 \\
    Random & 20.00 & 30.02 & 41.11 \\
    \bottomrule
    \end{tabular}
  \label{tab:topkmetrics}
\end{table}
\begin{table}[t]
\footnotesize
  \centering
    \caption{Evaluation of the CE Model's Discriminative Performance}
  \begin{tabular}{c|c|c|c|c}
    \toprule
    Accuracy & Precision & Recall & F1-score & AUC-ROC \\
    \hline
    90.56\% & 92.02\% & 90.56\% & 90.91\% & 96.02\% \\
    \bottomrule
  \end{tabular}
  \label{tab:metrics}
  \vspace{-1.0em}
\end{table}

\begin{table}[t]
\footnotesize
  \centering
    \caption{Top-$k$ Evaluation Results for Assessing Model Generalization.}
    \begin{tabular}{c|c|c|c}
    \toprule
    \textbf{Top-$k$} & \textbf{Top-$1$} & \textbf{Top-$3$} & \textbf{Top-$5$} \\
    \textbf{Method} & Success \% & Success \% &  Success \% \\
    \hline
    CE Model & 73.33 & 89.44 & 91.11  \\
    Random  & 18.89 & 33.33 & 45.56  \\
    \bottomrule
  \end{tabular}
  \label{tab:overallresults}
  \vspace{-1.0em}
\end{table}

\begin{figure}[t]
  \centering
  \includegraphics[trim={0 .1in 0 0}, clip, width=0.43\textwidth]{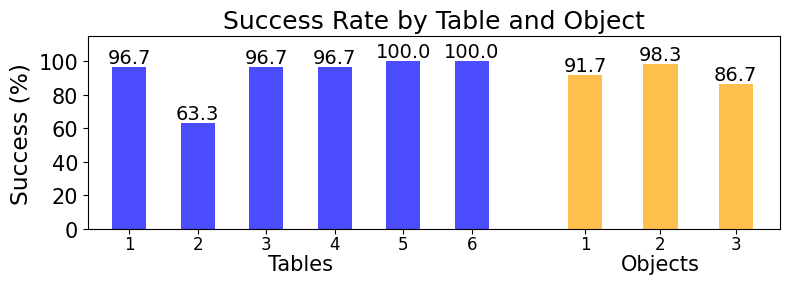}
  \caption{Object transport success rates by tables and objects.}
  \label{fig:successrates}
  \vspace{-.5em}
\end{figure}

The CE model is evaluated based on its ability to recommend feasible grasp configuration pairs among the top-$1$, top-$3$, and top-$5$ ranked candidates. Table~\ref{tab:overallresults} reports the overall success rates, including a random selection policy for reference. The results show that the proposed framework performs robustly even at small $k$, achieving a $91.11\%$ success rate with top-5 predictions---demonstrating strong generalization to novel objects and unseen environments. Figure~\ref{fig:successrates} further breaks down the success rates by object and table location, based on the top-5 selections. Among the test objects, Object 2 (triangular) achieved the maximum success rate at $98.3\%$, followed by Object 1 (T-shaped panel) at $91.7\%$, and Object 3 (asymmetric panel) at $86.7\%$. The average numbers of grasp configurations for Objects 1, 2, and 3 were $247$, $297$, and $267$, respectively.

\begin{figure}[t]
  \centering
  \includegraphics[trim={0 .05in 0 0}, clip, width=0.43\textwidth]{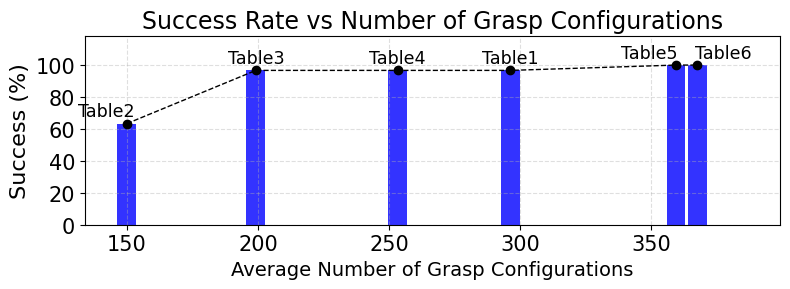}
  \caption{Object transport success rates in relation to the number of candidate grasp configurations across different tables.}
  \label{fig:evaluation_numgrasps_success}
  \vspace{-1.0em}
\end{figure}

Performance also varied across table locations. Tables 5 and 6 achieved perfect success rates of $100\%$, with average numbers of valid grasp configurations at 359 and 367, respectively. Tables 1, 3, and 4 exhibited a slightly lower success rate of $96.7\%$, with an average of $296$, $199$, and $253$ valid grasp configurations, respectively.  In contrast, Table 2 had the lowest success rate at $63.3\%$, with only $149$ valid grasp configurations on average. Most failures were concentrated around Table~2, where the object was tightly surrounded by obstacles (a wall and an adjacent table), severely restricting feasible grasp options. These results, as illustrated in Fig.~\ref{fig:evaluation_numgrasps_success}, suggest a positive correlation between the success rate and the number of candidate grasp configurations.

In summary, object geometry and grasping flexibility are key factors influencing performance. Object 3 (asymmetric panel) exhibited the lowest success rate, despite having more graspable elements, due to its complex shape that complicates trajectory planning. In contrast, Object 1 (T-shaped panel) performed better, though failures were more common in scenarios with geometric occlusions that limited grasping options. Object 2 achieved an almost perfect success rate, benefiting from its symmetric design and well-distributed grasp points, which allowed robust performance even in constrained environments such as Tables 2, 3, and 4.

\subsection{Evaluation on Physical Robot Platforms}

\begin{figure*}[t]
    \centering
    \makebox[\textwidth][c]{%
        \includegraphics[width=.16\textwidth]{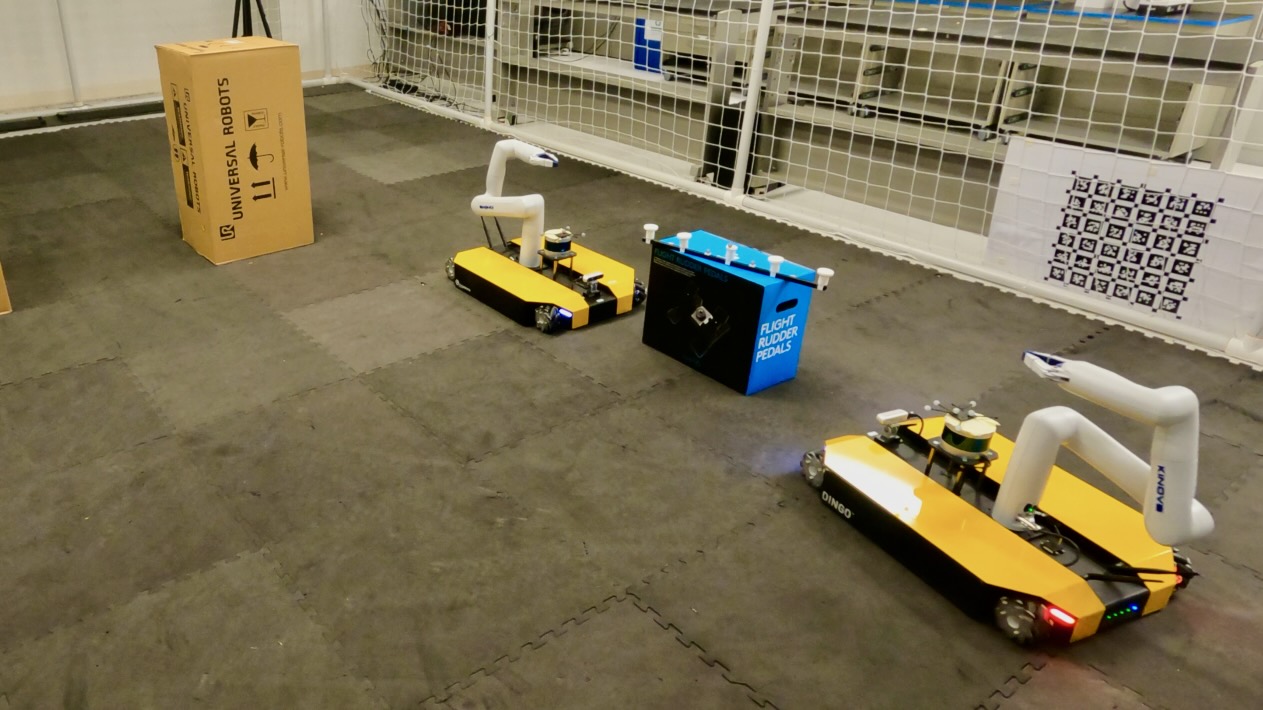}%
        \includegraphics[width=.16\textwidth]{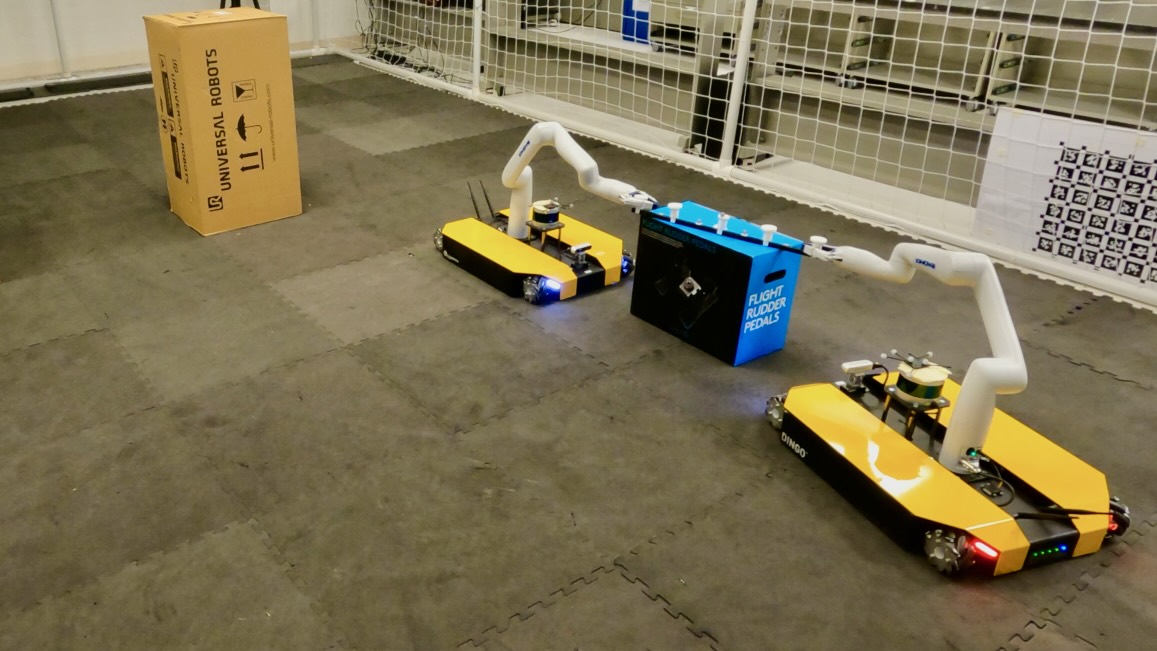}%
        \includegraphics[width=.16\textwidth]{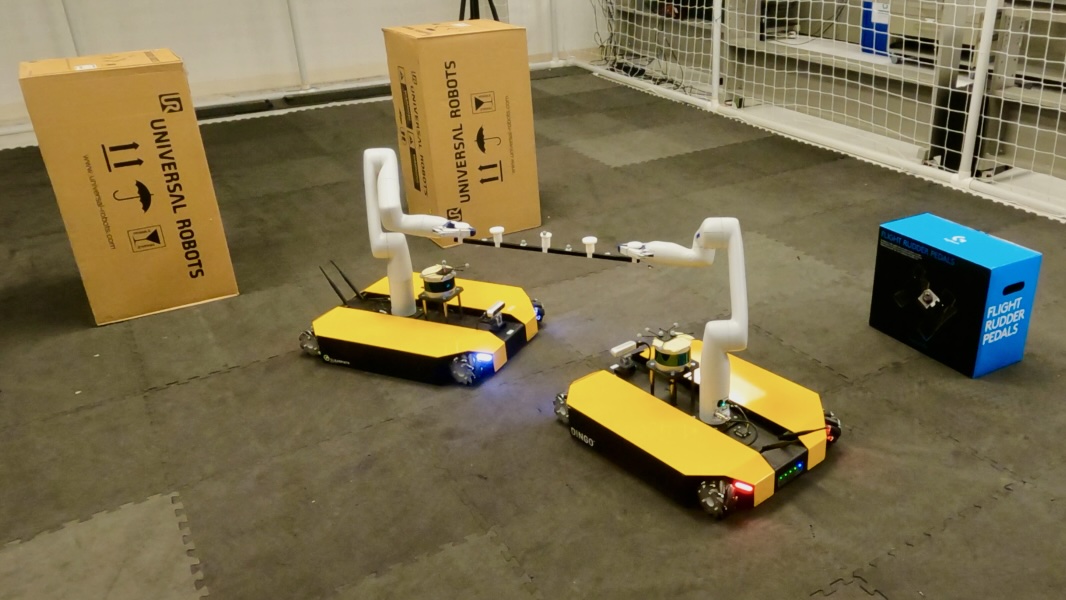}%
        \includegraphics[width=.16\textwidth]{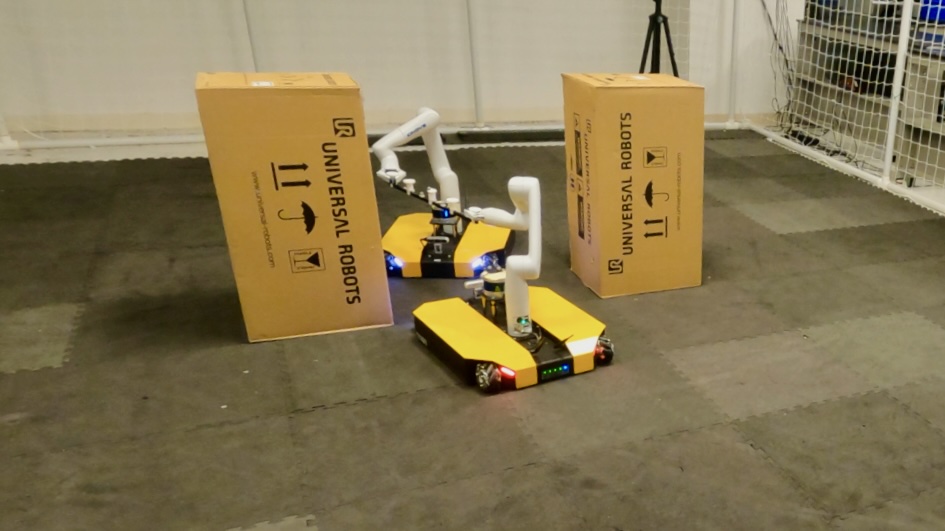}%
        \includegraphics[width=.16\textwidth]{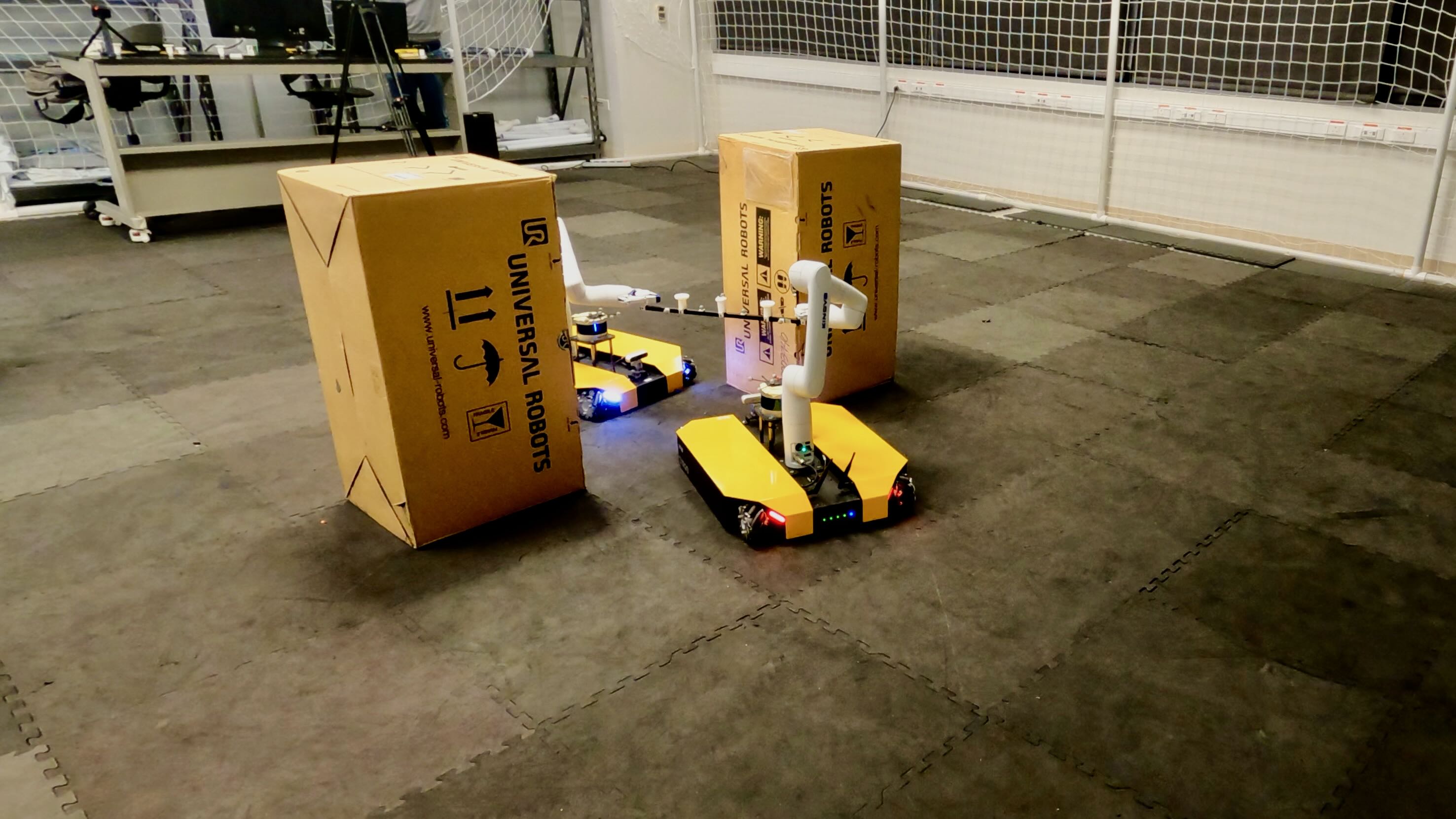}%
        \includegraphics[width=.16\textwidth]{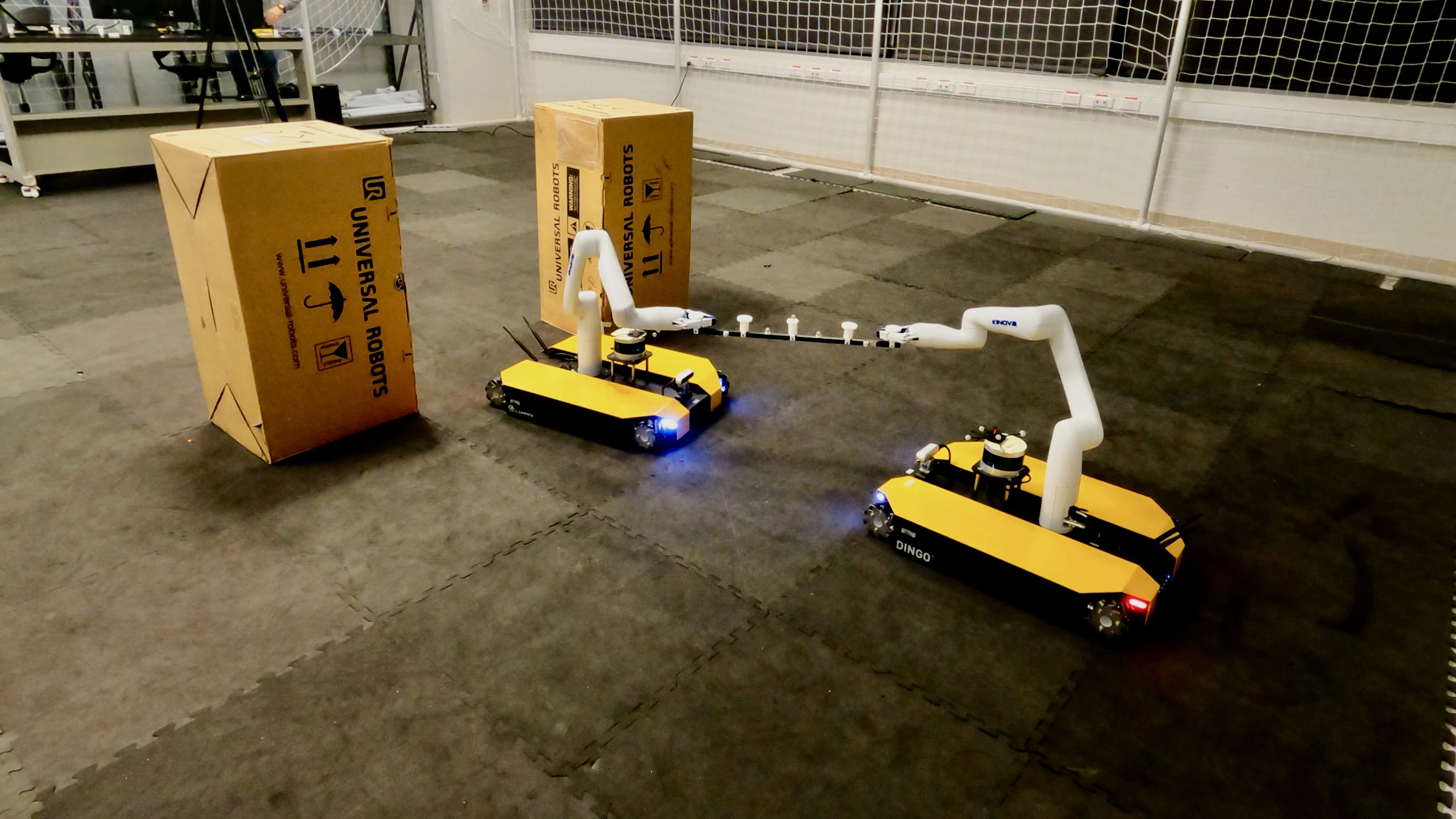}%
    }%
    \\[2pt]
    \makebox[\textwidth][c]{%
        \includegraphics[width=.16\textwidth]{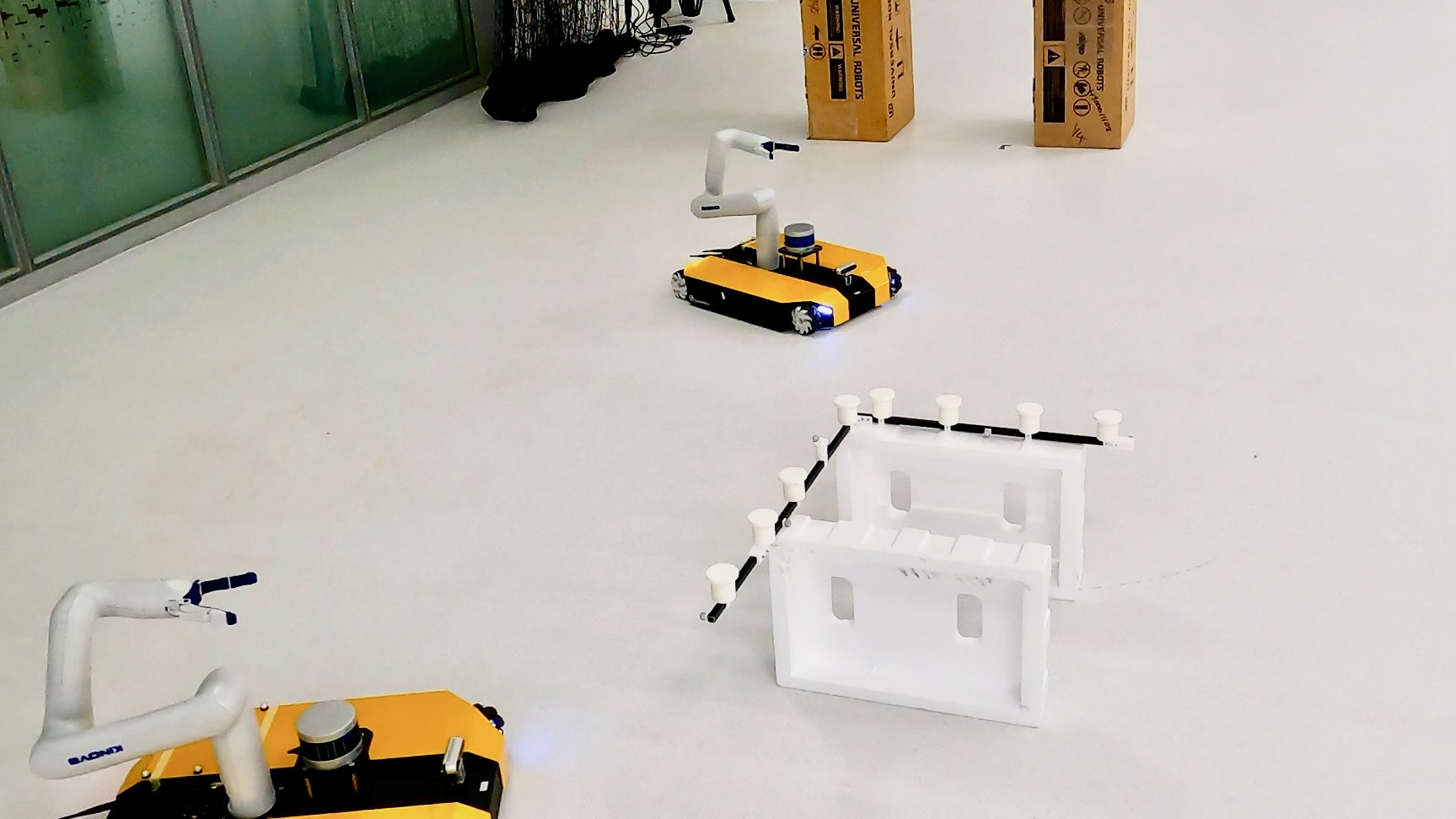}%
        \includegraphics[width=.16\textwidth]{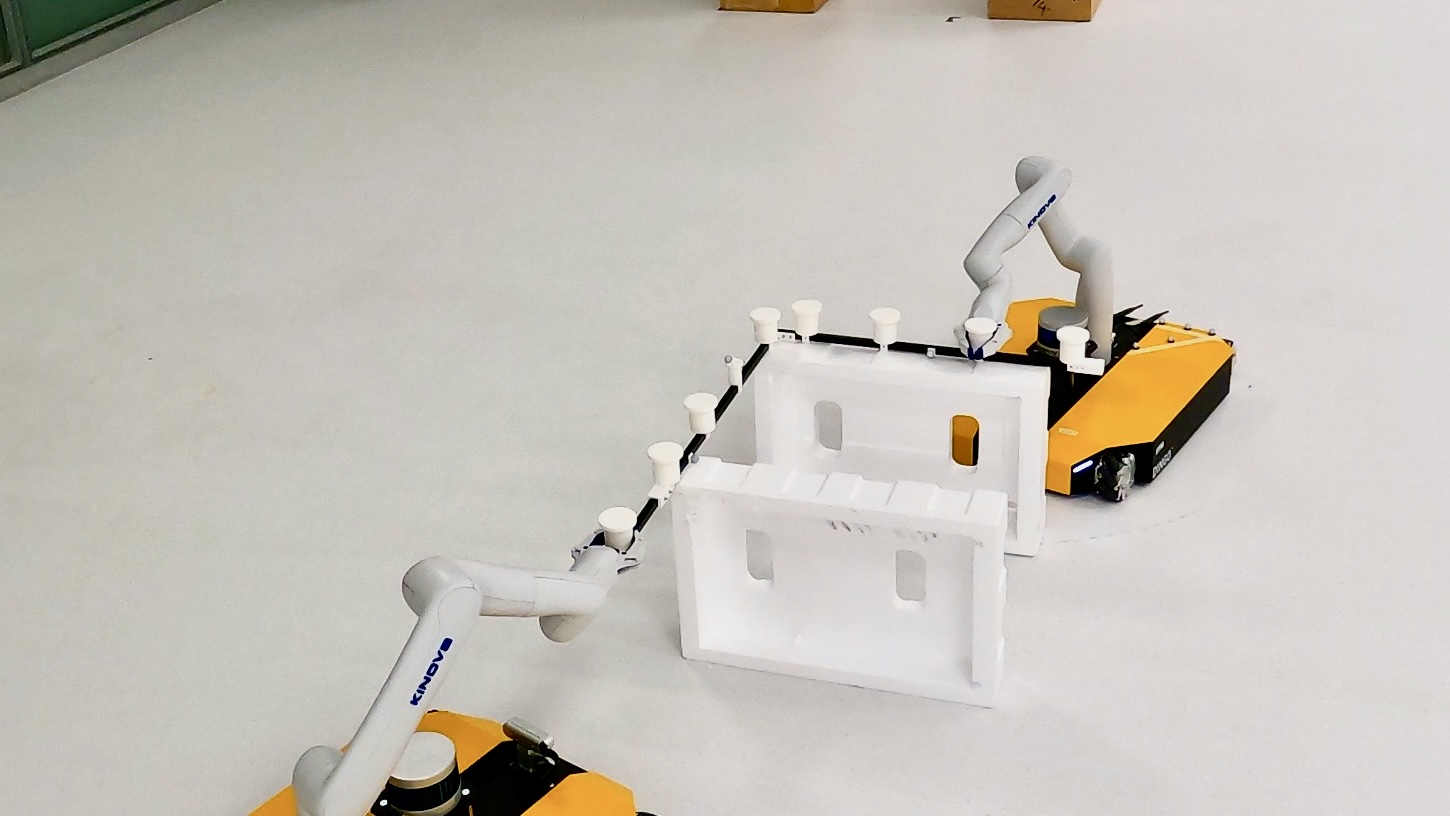}%
        \includegraphics[width=.16\textwidth]{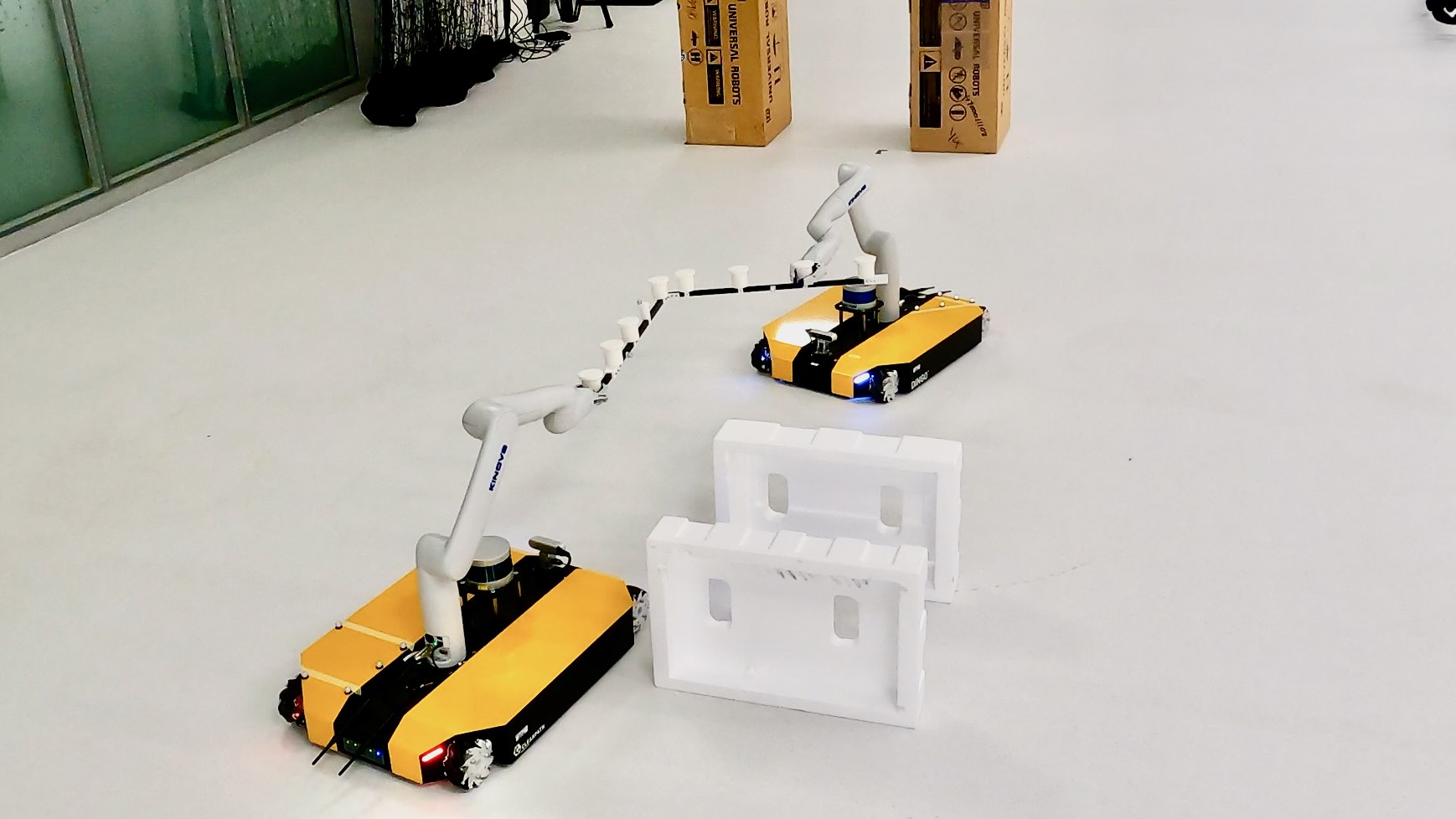}%
        \includegraphics[width=.16\textwidth]{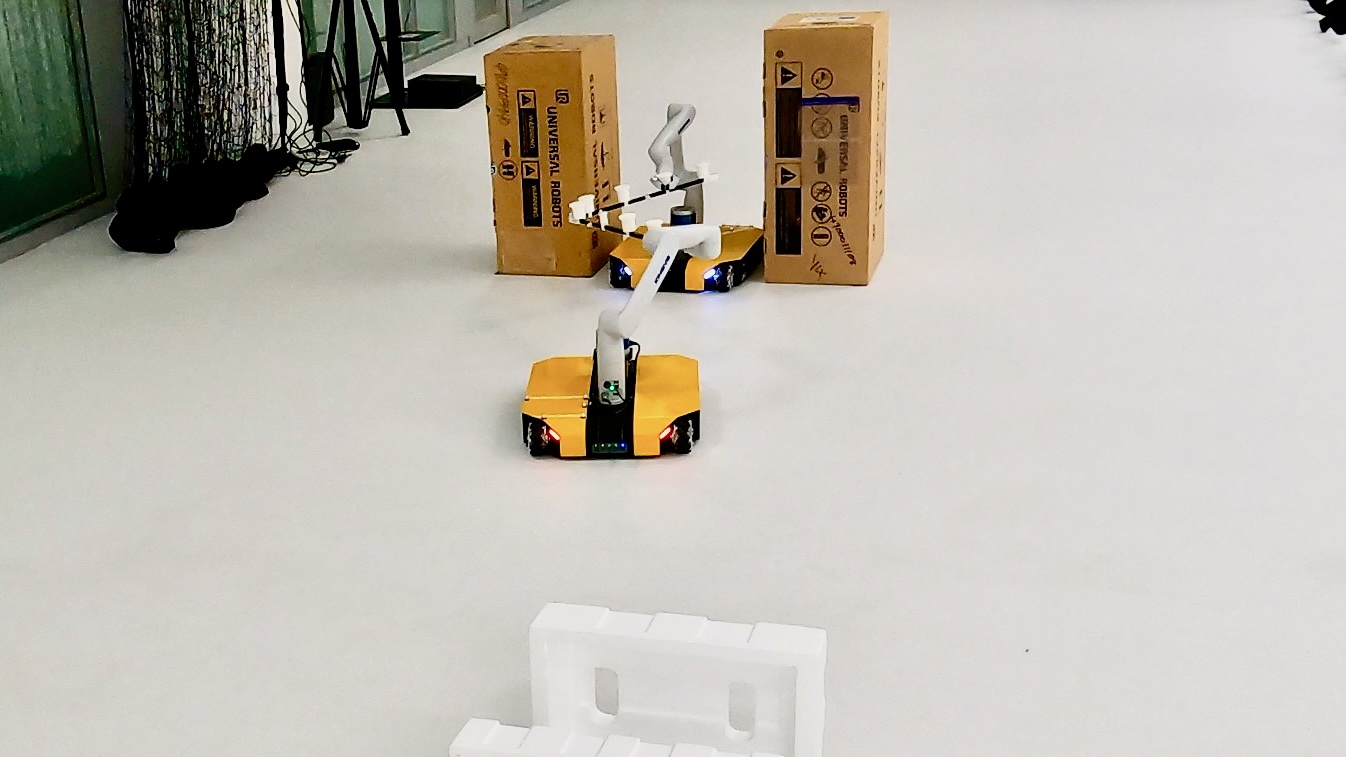}%
        \includegraphics[width=.16\textwidth]{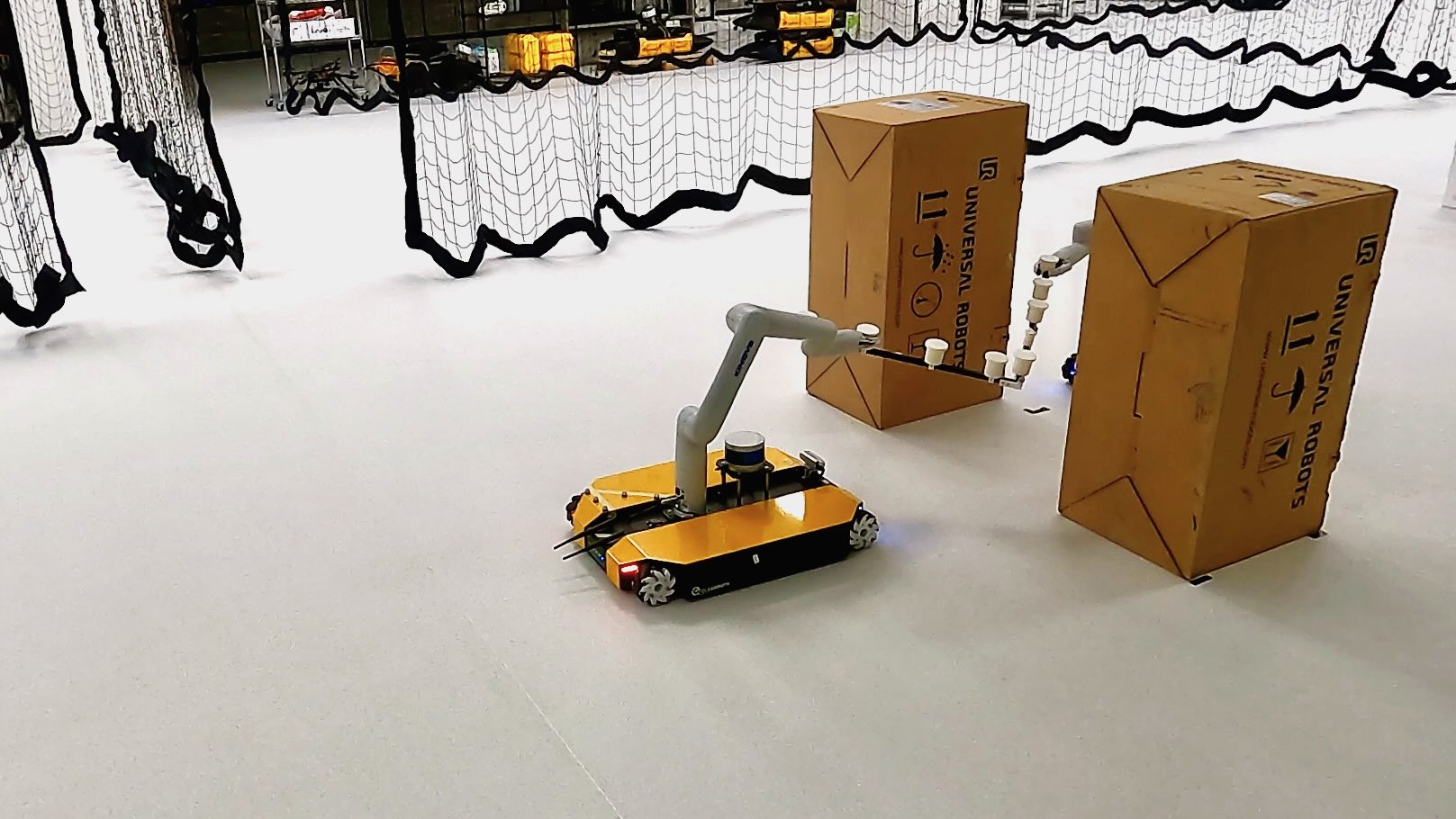}%
        \includegraphics[width=.16\textwidth]{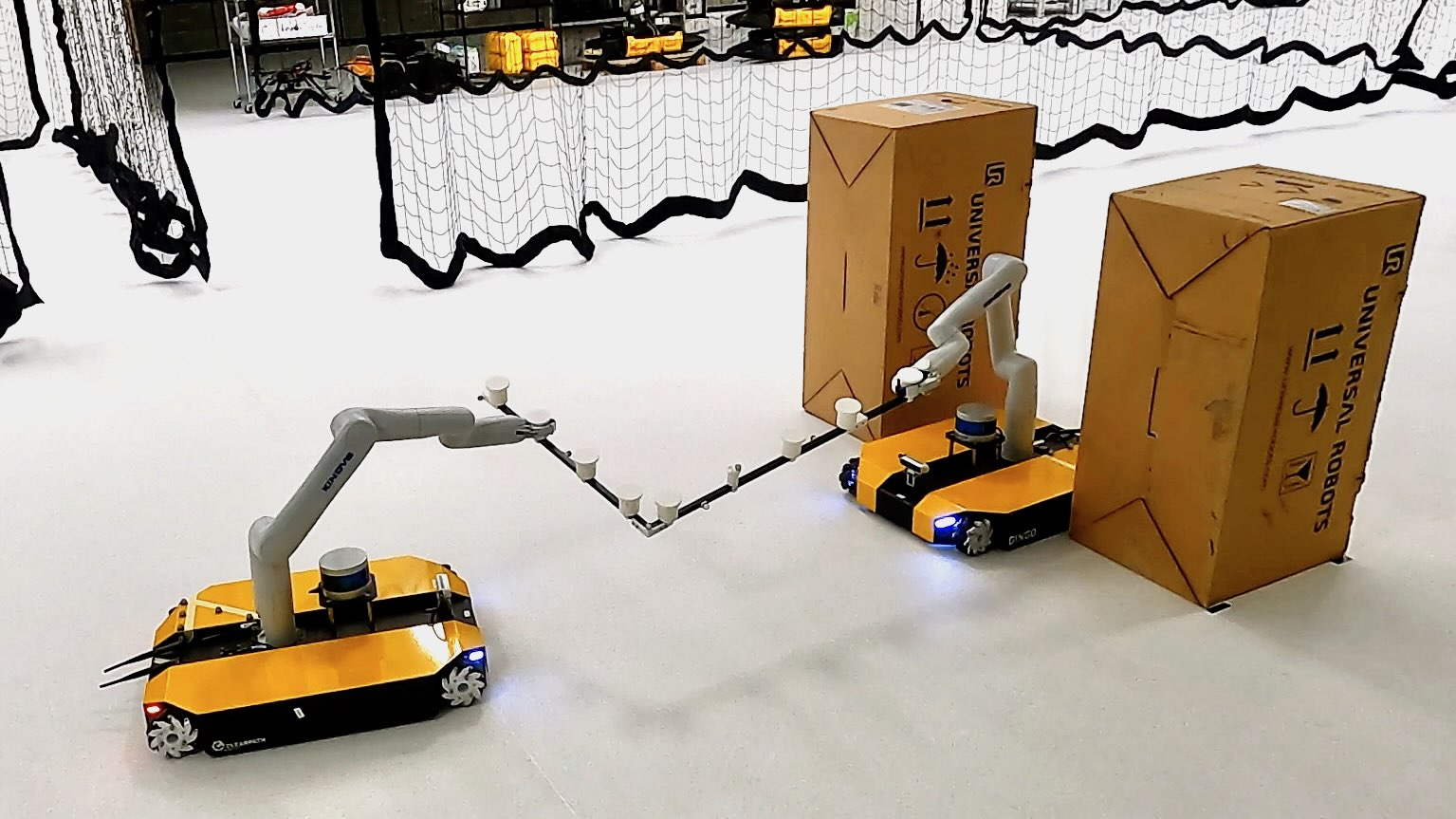}%
    }%
    \caption{Cooperative object transport using two \textit{Clearpath Dingo-O} mobile manipulators, each equipped
    with a \textit{Kinova Gen3-lite} arm.  
    \textbf{Top row:} The robots transport a straight bar through a narrow corridor connecting two areas. 
    \textbf{Bottom row:} The robots carry an L-shaped payload. Each Dingo-O platform measures $51\,\mathrm{cm}$ in width and $68\, \mathrm{cm}$ in length, while the corridor dimensions are $65\, \mathrm{cm}$ in width and $50\,\mathrm{cm}$ in length.}
    \label{fig:dingo_experiments_labs}
    \vspace{-1.0em}
\end{figure*}

We employed two \textit{Clearpath Dingo-O} omnidirectional mobile bases, each mounted with a \textit{Kinova Gen3-lite} manipulator. Robot and object poses were tracked using a motion capture system. The test environment, shown in Fig.~\ref{fig:multirobot_intro}, consists of two open areas connected by a narrow passage ($65\,\mathrm{cm} \times 50\,\mathrm{cm}$), allowing only one robot to pass at a time. An object is initially placed on a table in the first area, and the robots must transport it to a target location in the second area. We evaluated two object types: a straight bar ($63\,\mathrm{cm}$) and an L-shaped panel ($85\,\mathrm{cm} \times 60\,\mathrm{cm}$). In each trial, the CE model selected the grasp configuration with the highest affinity score, while the trajectory planner computed feasible paths for successful transport. These trajectories were then executed by each robot’s controller for navigation. Figure~\ref{fig:dingo_experiments_labs} shows snapshots of the experiments: the top row depicts transport of the bar, and the bottom row shows the L-shaped object being carried through the passage.

\section{Conclusions} \label{sec:conclusions}

This paper presented a framework for multi-robot decision-making in collective object transport. Central to the framework is the Conditional Embedding (CE) model, which maps information about candidate grasp configurations, object geometry, and environmental context into an embedding space. From this space, an affinity matrix is constructed and used to identify grasp configuration pairs that support effective execution of transport tasks. Evaluation results demonstrate the framework's ability to reliably identify feasible configurations, achieving high success rates across diverse scenarios. Moreover, its strong generalization to novel objects highlights the proposed framework's practical potential for multi-robot object transport applications. As future research, we aim to incorporate object regrasping to address the low success rates observed in scenarios with limited grasp configurations. For example, the robots could first move the object to a more open space, thereby increasing the number of feasible grasp candidates before regrasping it. In addition, extending the framework to scenarios involving more than two robots is also planned as future work.

\appendix

\subsection{Sampling Grasp Configuration Candidates} \label{sec:grasping_candidate_sampling} 
To grasp an object, each robot must move to a base position $(x_r, y_r)$ from which its end-effector can reach a designated grasp point $(x_g, y_g)$ selected from the predefined set of grasp points $\mathbb{F}$ on the object. For each $(x_g, y_g) \in \mathbb{F}$, we generate candidate base positions $(x_r, y_r)$ by uniformly sampling 60 points along a circle of radius $r$ centered at $(x_g, y_g)$. The sampling radius is set to $r = 0.55\,m$, which corresponds to an operationally effective reach distance from the center of the Dingo mobile base. This value falls within the robot arm’s minimum and maximum reach limits ($0.35\,m$ and $0.7\,m$, respectively), ensuring that any sampled base position $(x_r, y_r)$ allows the end-effector to reach the target grasp point $(x_g, y_g)$ through forward extension.

All sampled positions are then filtered to retain only those from which the manipulator can reach the grasp point $(x_g, y_g)$ without self-collision or collision with the tables and the object, and located within the free space $\mathbb M_{\text{free}}$. This process yields a set $\mathbb G$ of grasp configurations, where each configuration is represented as $G = (x_r, y_r, x_g, y_g)$.

\subsection{Multi-Robot and Object Trajectory Planning} \label{section:trajectory_planning}

We formulate the trajectory planning problem for multiple robots and an object as an optimization problem. Using IRIS-NP \cite{petersen_2023_iris}, the robot's configuration space in $\mathbb R^2$ is decomposed into $M$ overlapping convex polytopes $\{Q_r^{(1)}, \cdots, Q_r^{(M)}\}$ generated from $M$ manually selected ``seed'' points located in the collision-free space $\mathbb M_{\text{free}}$. The polytopes are constructed such that $Q_r^{(1)}$ contains the initial positions of the two robots, $Q_r^{(M)}$ contains their destination, and each consecutive pair $Q_r^{(i)}$ and $Q_r^{(i+1)}$ has a nonempty intersection.

Similarly, the object’s configuration space in $\mathbb{R}^2 \times [-\pi, \pi)$---representing its planar position and yaw orientation---is decomposed into a set of convex polytopes $\{Q_o^{(1)}, \cdots, Q_o^{(M)}\}$ using IRIS-NP. This decomposition leverages the same $M$ seed positions used for the robot’s configuration space, with each seed augmented by a carefully chosen orientation value from $[-\pi, \pi)$. The seeds are selected to lie entirely within the collision-free region of the extended configuration space. The polytopes are constructed such that $Q_o^{(1)}$ contains the object's initial position, $Q_o^{(M)}$ contains its destination, and each consecutive pair of polytopes, $Q_o^{(i)}$ and $Q_o^{(i+1)}$, is constructed to ensure a nonempty intersection.\footnote{In applying IRIS-NP seed selection is performed manually and judiciously, as suggested in \cite{marcucci_2022_convex_planning}, to ensure that the configuration spaces of both the robot and the object are decomposed in a way that supports efficient trajectory optimization.}

The set $Q = \bigcup_{k=1}^M Q^{(k)}$, where each region $Q^{(k)} = (Q_r^{(k)}, Q_o^{(k)})$, defines the joint feasible space of robot and object configurations. A trajectory $q: [0, T] \to Q$, comprising the positions of the two robots $\mathrm x_{r_1}(t), \mathrm x_{r_2}(t) \in \mathbb R^2$ and the position and yaw orientation of the object $\mathrm x_{o}(t) \in \mathbb R^2, \theta_{o}(t) \in [-\pi,\pi)$, is then planned to ensure collision-free motion while satisfying the kinematic constraints of both the robots and the object. Following the method in \cite{marcucci_2022_convex_planning}, we segment the trajectory $q(t), \, t \in [0, T]$ based on its inclusion in regions $Q^{(k)}$, and parameterize each segment as a Bézier curve constrained to lie within its corresponding region $Q^{(k)}$. Each segment ${q_k}(t), \, t \in [t_k, t_{k+1})$ is defined by four control points $\mathbf{c}_0^{(k)}, \cdots, \mathbf{c}_3^{(k)}$. Since resulting Bézier curves lie within the convex hull of their control points, ensuring $\mathbf{c}_0^{(k)}, \cdots, \mathbf{c}_3^{(k)} \in Q^{(k)}$ guarantees that the entire segment remains within the collision-free region. The control points $\mathbf{c}_0^{(k)}, \cdots, \mathbf{c}_3^{(k)} \in Q^{(k)}, \,  k = 1, \cdots, M$ defining the entire trajectory from the initial state $q_0$ to the final state $q_T$ are obtained by solving the following optimization problem:

{\small
\begin{subequations}
    \begin{align}
    \min \quad &
    L(\dot q, T)\!+\!V(\dot q, T)\!+\!S(\ddot q, T)\!+\!w_F F(\dot q, T) 
    \label{eq:optimization_problem} \\
    \text{s.t.} \quad 
    & q(0) = q_0, \quad q(T) = q_T \notag \\ 
    & \dot{q}(0) = 0, \quad \dot{q}(T) = 0 \notag \\ 
    & ||\mathrm x_{r_i} (t) \!-\! P_{G}(t) \bar{\mathrm{x}}_{g_i}||_2 \!\geq\! r_{\text{min}}, \, t \in [0, T], \, i = 1,2 \label{eq:const_minmax_radius_1} \\
    & ||\mathrm x_{r_i} (t) \!-\! P_{G}(t) \bar{\mathrm{x}}_{g_i}||_2 \!\leq\! r_{\text{max}}, \, t \in [0, T], \, i = 1,2  \label{eq:const_minmax_radius_2} \\
    & ||\mathrm x_{r_i} (t) \!-\! P_{G}(t) \bar{\mathrm{x}}_{r_i}||_2 \!\leq\! r_{\text{collision},i}, \, t \in [0, T], \, i = 1,2 \label{eq:const_formation_radius} \\
    & \mathbf c_{3}^{(k)} - \mathbf c_{2}^{(k)}  = \mathbf c_{1}^{(k+1)} - \mathbf c_{0}^{(k+1)}, \label{eq:const_continuity_1} \\ 
    & \mathbf c_{3}^{(k)} = \mathbf c_{0}^{(k+1)}, ~ k = 1, \cdots, M-1 \label{eq:const_continuity_2} 
\end{align}
\end{subequations}
}
where
{\small
\begin{align*}
    L(\dot{q}, T) &= \textstyle\int_{0}^{T} \|\dot{q}(t)\|_2 \, \mathrm dt, \\
    V(\dot{q},T) &= \textstyle \int_{0}^{T} \|\dot{q}(t)\|_2^{2}\, \mathrm dt, \quad 
    S(\ddot{q},T) = \textstyle \int_{0}^{T} \|\ddot{q}(t)\|_2^{2}\, \mathrm dt, \\
    F(\dot{q}, T) &= \textstyle \sum_{i=1}^{2} \int_{0}^{T} \|\dot{\mathrm{x}}_{r_i}(t) - (\dot{\mathrm{x}}_{o}(t) + \dot{\theta}_{o}(t) \times \bar{\mathrm{x}}_{r_i})\|_2^{2} \, \mathrm dt.
\end{align*}
}

The objective function in \eqref{eq:optimization_problem} is composed of four terms: the trajectory length $L(q,T)$, the velocity penalty $V(\dot{q},T)$, the smoothness term $S(\ddot{q},T)$, and the formation consistency term $F(\dot{q}, T)$. The last term $F(\dot{q}, T)$ measures the deviation between each robot $i$'s velocity $\dot{\mathrm{x}}_{r_i}(t)$ and its expected rigid-body velocity, given by $\dot{\mathrm{x}}_{o}(t) + \dot{\theta}_{o}(t) \times \bar{\mathrm{x}}_{r_i}$. Here, $\bar{\mathrm{x}}_{r_i}$ denotes the relative position of the robot from the center of the object, based on the selected grasp configuration $G$, and is treated as a point rigidly attached to the object with translational velocity $\dot{\mathrm{x}}_{o}(t)$ and angular velocity $\dot{\theta}_{o}(t)$. This term penalizes deviations from ideal rigid-body behavior among the robots and the object. To control the influence of this term, we introduce a weight factor $w_F$ (set to $20.0$ in our implementation) that scales the contribution of $F(\dot{q}, T)$ in the overall objective.

To ensure feasible and coordinated manipulation, we impose three constraints. Firstly, each robot~$i$ must remain within a radial annulus defined by a minimum radius $r_{\text{min}}=0.35$ and a maximum radius $r_{\text{max}} = 0.7$, centered at its current grasp point $P_G(t) \bar{\mathrm{x}}_{g_i}$ (constraints~\eqref{eq:const_minmax_radius_1}, \eqref{eq:const_minmax_radius_2}), where $P_G(t)$ is a time-varying transformation matrix aligning the initial grasp point $\bar{\mathrm x}_{g_i}$ with the current grasp point $P_G(t) \bar{\mathrm{x}}_{g_i}$.
Secondly, the formation constraint~\eqref{eq:const_formation_radius} ensures that each robot $i$ remains within a distance $r_{\text{collision}, i}$ of its nominal base position $P_G(t) \bar{\mathrm{x}}_{r_i}$, where $r_{\text{collision}, i}$ is selected to as the maximum distance that guarantees collision-free operation between each robot~$i$ and the object, given a grasp configuration. 
Finally, continuity of the trajectory is enforced by ensuring $q(t)$ is continuously differentiable, as specified in constraints~\eqref{eq:const_continuity_1}, \eqref{eq:const_continuity_2}.


\balance
\bibliographystyle{IEEEtran} 
\bibliography{References}

\end{document}